%% file: 0094.tex
\documentclass[10pt,twocolumn,letterpaper]{article}

\usepackage{wacv}
\pdfoutput=1

\usepackage[accsupp]{axessibility}
\usepackage{times}
\usepackage{epsfig}
\usepackage{graphicx}
\usepackage{amsmath}
\usepackage{amssymb}
\usepackage{bm}
\usepackage[norelsize, linesnumbered, ruled, lined, boxed, commentsnumbered]{algorithm2e}
\usepackage{caption}
\usepackage{subcaption}
\usepackage{comment}
\usepackage{enumitem}
\usepackage{dirtytalk}
\usepackage{soul}
\usepackage{authblk}
\usepackage{multirow}

\def \gtSeg{\mathcal{G}} % ground truth segmentation of an image 
\def \spSeg{\mathcal{L}} % superpixel segmentation of an image
\def \ncluster{K} % number of clusters
\def \pixel{\bm{p}}

\newcommand{\nclass}{C}
\newcommand{\gausssim}{g}
\newcommand{\ournet}{DAL }
\newcommand{\graph}{\mathcal{G}}
\newcommand{\nodes}{\mathcal{V}}
\newcommand{\edges}{\mathcal{E}}

\def\wacvPaperID{94} % Enter the WACV Paper ID here

\wacvfinalcopy % *** Uncomment this line for the final submission

\ifwacvfinal
\usepackage[breaklinks=true,bookmarks=false]{hyperref}
\else
\usepackage[pagebackref=true,breaklinks=true,colorlinks,bookmarks=false]{hyperref}
\fi

\begin{document}

%%%%%%%%% TITLE
\title{HERS Superpixels: Deep Affinity Learning for Hierarchical Entropy Rate Segmentation}

%%%%%% Options for titles:
%  HERS Superpixels: Deep Affinity Learning for Accelerated  Hierarchical Entropy Rate  Segmentation

% \author{First Author\\
% 	Institution1\\
% 	Institution1 address\\
% 	{\tt\small firstauthor@i1.org}
% 	% For a paper whose authors are all at the same institution,
% 	% omit the following lines up until the closing ``}''.
% 	% Additional authors and addresses can be added with ``\and'',
% 	% just like the second author.
% 	% To save space, use either the email address or home page, not both
% 	\and
% 	Second Author\\
% 	Institution2\\
% 	First line of institution2 address\\
% 	{\tt\small secondauthor@i2.org}
% }
\author[1]{Hankui Peng}
\author[1]{Angelica I. Aviles-Rivero}
\author[1]{Carola-Bibiane Sch\"{o}nlieb}

\affil[1]{DAMTP, University of Cambridge. \authorcr
	\{\tt hp467, ai323, cbs31\}@cam.ac.uk}

\maketitle
\thispagestyle{empty}

%%%%%%%%% ABSTRACT
\begin{abstract}
	Superpixels serve as a powerful preprocessing tool in numerous computer vision tasks. By using superpixel representation, the number of image primitives can be largely reduced by orders of magnitudes. 
	%
	%The majority of superpixel techniques are developed based on handcrafted features. 
	With the rise of deep learning in recent years, a few works have attempted to feed deeply learned features / graphs into existing classical superpixel techniques.	
%	The majority of superpixel methods use handcrafted features, which usually do not translate well into strong adherence to object boundaries. A few recent superpixel methods have introduced deep learning into the superpixel segmentation process. 
	However, none of them are able to produce superpixels in near real-time, which is crucial to the applicability of superpixels in practice.
	In this work, we propose a two-stage graph-based framework for superpixel segmentation. In the first stage, we introduce an efficient Deep Affinity Learning (DAL) network that learns pairwise pixel affinities by aggregating multi-scale information. In the second stage, we propose a highly efficient superpixel method called Hierarchical Entropy Rate Segmentation (HERS). Using the learned affinities from the first stage, HERS builds a hierarchical tree structure that can produce any number of highly adaptive superpixels instantaneously. 
	%\Angie{In this work, we propose a framework for superpixel segmentation that combines a neural network architecture with a highly efficient graph-based segmentation technique. Firstly, we introduce an efficient network, which learn pixels affinities by aggregating multi-scale information. Secondly, we introduce a graphical superpixel technique, called HERS, which }
	%In this paper, we propose a framework that consists of a graph affinity learning component and a highly efficient Hierarchical Entropy Rate Segmentation (HERS) algorithm. 
	We demonstrate, through visual and numerical experiments, 
	%\st{Experiments on standard superpixel benchmark datasets demonstrate} 
	the effectiveness and efficiency of our method compared to various state-of-the-art superpixel methods.~\footnote{The code is available at: \url{https://github.com/hankuipeng/DAL-HERS}}
\end{abstract}

%%%%%%%%% BODY TEXT
\section{Introduction}
Superpixel segmentation is the task of partitioning an image into meaningful regions, within which the pixels share similar qualities such as colour, texture, or other low-level features. It is a powerful preprocessing tool for various computer vision tasks. For example, image classification~\cite{fang2015spectral,sellars2020superpixel}, optical flow \cite{menze2015object,liu2019selflow}, object tracking~\cite{wang2011superpixel,yang2014robust} and semantic segmentation~\cite{kwak2017weakly, zhao2018improved}. Superpixels offer more computationally digestible input data representations than the conventional pixel-level format. By using superpixels, one can substantially reduce the number of image primitives, whilst highlighting the discriminative information~\cite{achanta2012slic}. 

The aforementioned advantages have encouraged the fast advancement of superpixel segmentation techniques (e.g.~\cite{levinshtein2009turbopixels,achanta2012slic,liu2011entropy,li2015superpixel,maierhofer2018peekaboo,vedaldi2008quick}) since the seminal work of Ren and Malik~\cite{ren2003learning}. For a superpixel segmentation method to be useful, it should be computationally efficient and preserve the structure of the objects well (i.e.\ adhere to the object boundaries). 
%Although the body of literature has reported promising results, the majority of existing works rely on hand-crafted features. This design choice limits the generalisability of superpixels across various types of scenes.% , which results in unsuccessfully preserving fine details and structures in the image. 
With the advent of deep learning, it becomes possible to explore more flexible representations in various superpixel segmentation techniques. 
%However, unlike other tasks, where the adoption of deep learning has been widely applied, superpixels have been only recently used along with deep nets. 
However, to the best of our knowledge, there are only a few attempts to employ deep networks for superpixels (e.g.~\cite{tu2018learning,jampani2018superpixel,yang2020superpixel}). \\
\indent There are several reasons for this limited adoption. Firstly, the conventional convolution operation in a neural network is designed to work efficiently with a regular image grid, whereas superpixel segmentations naturally give rise to irregular grids. 
%However, this is not the case for irregular superpixels grids. 
Secondly, there does not exist ground truth superpixels in an image, but rather a delineation of the  object structure. Thirdly, several existing superpixel techniques (e.g.~\cite{achanta2012slic, liu2011entropy}) are non-differentiable, due to the nearest neighbour assignment used in the pixel-superpixel association. This imposes a challenge in the network training process which would be end-to-end trainable otherwise. \\
\indent
%% clustering-based (learn the pixel features)
So far, there are only a few works that have addressed these challenges in integrating deep networks within superpixel methods. % An alternative to previous challenges is to 
% approach 1
One possible approach is to redesign a network architecture to allow the computation of irregular superpixel grids, e.g.~\cite{gadde2016superpixel,suzuki2018superpixel,yang2020superpixel}. This is still challenging, particularly, if one wishes to integrate subsequent tasks within a single learning process. 
% approach 2 
An alternative approach is to decouple the superpixel segmentation process from the deep network training process. For example, one can use a network to extract pixel-level features and then feed them into a superpixel segmentation method~\cite{jampani2018superpixel}. %, or use a network to learn pixel-wise affinities and use them as input to a superpixel segmentation method~\cite{liu2011entropy}.
%Whilst this option allows for the use of deep networks, one needs to rethink the superpixels principles while preserving their properties and performance. %This option is still challenging and far from being trivial. 
Most of these existing attempts are based on building upon a modified version of SLIC~\cite{achanta2012slic}, which is based on Lloyd's algorithm for $k$-means clustering, for its efficiency and simplicity. However, SLIC has a few limitations, such as the over-partition in homogeneous regions of an image and high computational cost in texture-rich regions. \\
\indent
%% graph-based (learn the affinities)
As an alternative to clustering-based techniques, a number of researchers have demonstrated the potential of graph-based superpixel techniques~\cite{ren2003learning, tu2018learning}. 
% challenges 
Under this framework, the superpixel segmentation task is translated into a graph partitioning one. The key challenge in this framework lies in obtaining a good partition of the graph. In addition, the problem of learning pixel-wise affinities from a standard deep network that can translate well into edge weights of a graph is not trivial. 
% advantages 
With that said, the graph-based framework provides several advantages such as computational tractability, natural representation of an image, and desirable tree structure.

\begin{figure}[ht!]
	\centering
	\begin{subfigure}{.49\linewidth}
		\includegraphics[width=\linewidth, height=.65\linewidth]{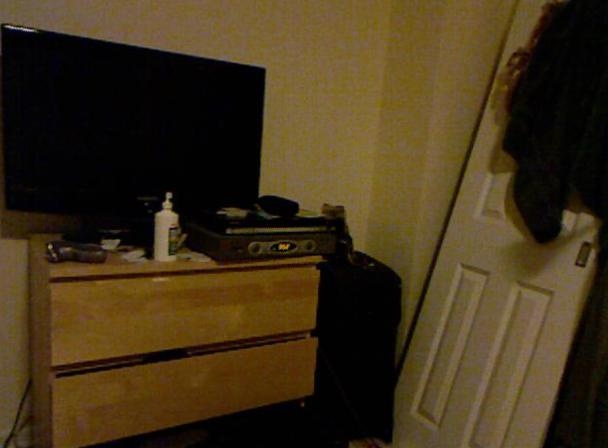}
		%\caption{Original image.}
	\end{subfigure}
	% 	\begin{subfigure}{.325\linewidth}
	% 		\includegraphics[width=\linewidth]{imgs/00061/00061_ours_merged.png}
	% 		%\caption{Merged image.}
	% 	\end{subfigure}
	\begin{subfigure}{.49\linewidth}
		\includegraphics[width=\linewidth, height=.65\linewidth]{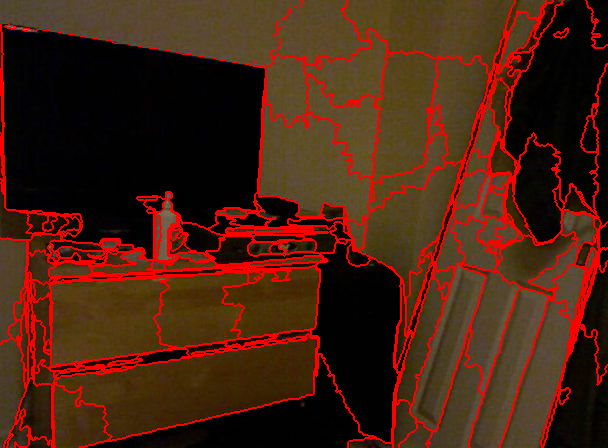}
		%\caption{Segmented image.}
	\end{subfigure}
	\caption{HERS produces superpixels that are highly adaptive to the homogeneous regions of an image. \textbf{Left}: the original image. \textbf{Right:} segmentation by DAL-HERS.} %\vspace{-0.3cm} 
	\label{fig:HERS}
	\vspace{-.6cm}
\end{figure}

%% paragraph for contributions
In this work, we present a two-stage graph-based superpixel segmentation framework, where an output example is displayed in Figure~\ref{fig:HERS}. 
In the first stage, we propose a deep network structure, Deep Affinity Learning (DAL), to learn boundary-aware pixel-wise affinities. In the second stage, we propose an efficient graph partitioning method called Hierarchical Entropy Rate Segmentation (HERS). HERS produces highly adaptive superpixels that adhere well to the object boundaries by maximising the entropy rate of the graph.
%a novel superpixel technique called \Angie{[]}. Our technique follows the graph based perspective, in which we seek to compute deeply learned affinities for a graph. Our proposal encourages adaptive superpixel sizes whilst enforcing balanced cluster sizes. \Angie{to update this part with strong motivation regarding the design after taking a global look at the paper.} 
To summarise, our main contributions are:
\vspace{-.5cm}
\begin{itemize}[noitemsep]
	\item We propose DAL network that learns and aggregates multi-scale information, produces boundary-aware affinities, and can be trained efficiently.  
	\item We propose an efficient superpixel segmentation algorithm called HERS that produces highly adaptive superpixels. It builds a hierarchical tree structure that allows instant generation of any number of superpixels. 
	\item We evaluate our proposal on superpixel benchmark datasets that contain a range of indoor and outdoor scenes. Extensive experimental results demonstrate its effectiveness (numerically and visually) and efficiency against various state-of-the-art classical and deep learning based superpixel methods.
	%	\item We demonstrate that our technique leads to a better partition of the image than existing techniques, and can readily compete with existing deep learning methods.  
\end{itemize}

%-------------------------------------------------------------------------
\section{Related Work}
Superpixel segmentation methods have been extensively studied in the literature, with the majority of existing techniques designed from the classical perspective. Most recently, a few works have reported the use of deep networks for superpixels. %In this section,
We review the existing techniques in turn.

\smallskip
\textbf{Classic techniques for superpixels.} Since the pioneering work of \cite{ren2003learning}, several techniques have been proposed  which can be  roughly divided into patch-based models~\cite{fu2014regularity,tang2012topology}, watershed techniques~\cite{gonfaus2010harmony, benesova2014fast, machairas2015waterpixels},  clustering-based approaches~\cite{achanta2012slic,li2015superpixel,liu2016manifold}
% , achanta2017superpixels,maierhofer2018peekaboo,zhang2021dynamic
and graph-based techniques~\cite{ren2003learning,felzenszwalb2004efficient,liu2011entropy}. 
% ,humayun2015middle
The last two categories are the most widely applied family of techniques, which we will cover in the rest of this section. 

One set of techniques have been proposed based on clustering principles for superpixels.  The most popular technique in this category is the Simple Linear Iterative Clustering (SLIC)~\cite{achanta2012slic}. This technique partitions a given image using a local version of the $k$-means algorithm. Whilst this technique offers simplicity and efficiency, it has several drawbacks. For example, it unnecessarily partitions uniform areas and computes unnecessary distances in dense areas. These issues motivated several improvements for $k$-means based techniques including reducing the number of distance calculations, improving the seeding initialisation and improving the feature representation~\cite{liu2016manifold,li2015superpixel,achanta2017superpixels,maierhofer2018peekaboo,zhang2011superpixels}. 

\begin{figure*}[t!]
	\begin{center}
		%\fbox{\rule{0pt}{2in} \rule{.9\linewidth}{0pt}}
		\includegraphics[width=\textwidth]{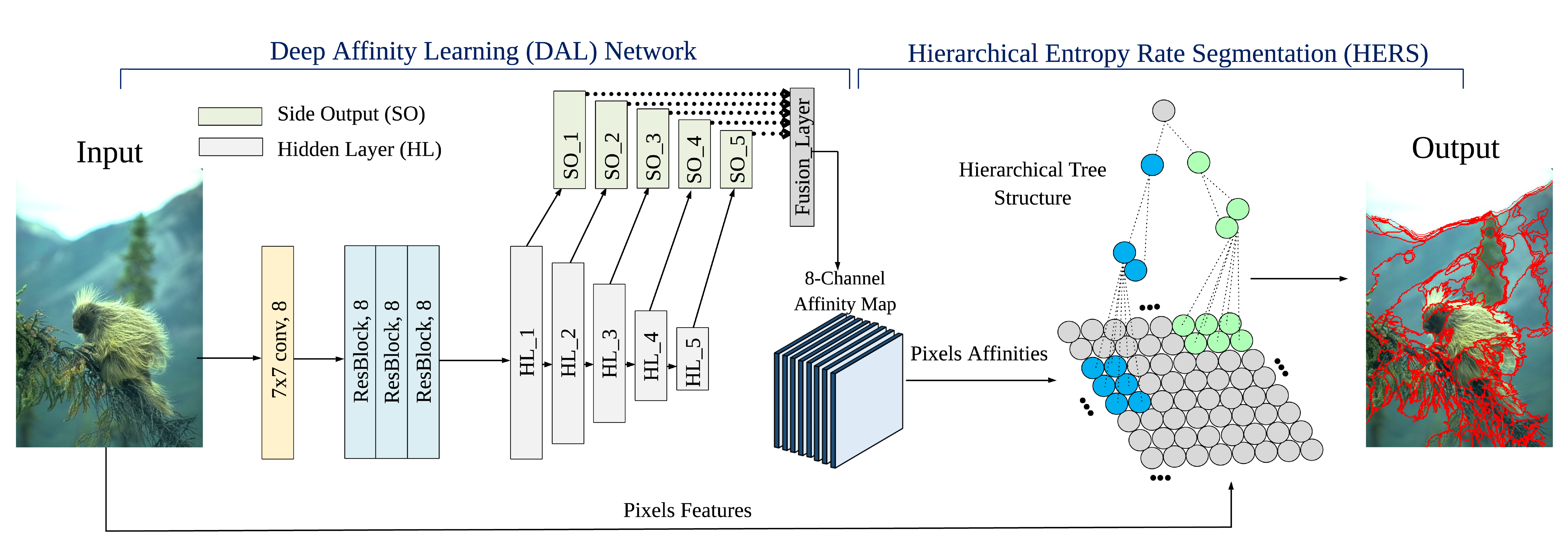}
	\end{center}
	\caption{Illustration of our proposed superpixel framework. The left side of the scheme shows the flow of our proposed DAL network, which learns an 8-channel pixel affinity map. The right side displays our proposed HERS algorithm, which constructs a hierarchical tree structure that allows any number of highly adaptive superpixels to be generated instantaneously.} %\vspace{-0.2cm}
	\label{fig:network_flow}
	% the network is made in lucid.app -> https://lucid.app/lucidchart/126aeae9-4726-4603-b913-f4be96715260/edit?page=_31#
\end{figure*}
Another set of techniques treat superpixel segmentation as a graph partitioning problem. 
The seminal paper of~\cite{ren2003learning} uses the Normalised Cuts algorithm, where superpixels are subgraphs that are obtained as a result of the partitioning based on edge similarities. This technique produced fair results, with boundary adherence being the main issue. 
Another graph based technique was presented in~\cite{felzenszwalb2004efficient}, where the main criterion for graph partitioning is based on the evidence for a boundary between two regions. 
%The evidence relies on the intensity difference firstly on the boundaries and secondly between neighbouring pixels. 
%This work inspired several other techniques such as the one in~\cite{wei2018superpixel}, where the weights on the graph are being adjusted iteratively. 
One of the most competitive graph-based techniques is Entropy Rate Superpixel Segmentation (ERS)~\cite{liu2011entropy}. It considers edge weights as transition probabilities of a random walk, and selects edges to form a number of subgraphs / superpixels that maximises the overall entropy rate of the partition. The problem is solved through a lazy greedy algorithm~\cite{nemhauser1978analysis}, which is not very computationally efficient. 
%Several previous techniques, e.g.~\cite{grady2006random,veksler2010superpixels,zhang2011superpixels,li2012segmentation}, 
% ,shen2014lazy,humayun2015middle
%serve as the basis for other graph-based techniques.

\smallskip
\textbf{Deep learning techniques for superpixels.} 
%A commonality of classical techniques for superpixels is the use of hand-crafted features, which limits their generalisability. Whilst this problem has been addressed through the use of deep networks in many other computer vision tasks, this is not the case for superpixel segmentation where only a few works have been reported that employ deep networks. 
Given the impressive results achieved by deep learning in recent years, a few recent works have introduced the idea of deeply learned features or edge affinities, in place of handcrafted ones, in the context of superpixel segmentation. 
%The application of deep learning has been  formally introduced in the context of superpixels 
In~\cite{jampani2018superpixel}, the authors proposed the SSN model that uses a deep network to extract pixel-level features, and then use the learned features as input to a soft version of SLIC~\cite{achanta2012slic}. The key idea is to enforce  soft-associations between pixels and superpixels to avoid the non-differentiability of SLIC. 
% critique of the method 
Although SSN extends the capability of SLIC through learned features, it produces superpixels which mimics that of SLIC, which are not very adaptive in sizes.
The authors of~\cite{tu2018learning} introduced a graph-based model SEAL, where the key idea is to learn the pixel affinities for superpixel segmentation. This work is built upon ERS~\cite{liu2011entropy} along with a segmentation-aware loss. Although it produces very impressive results quantitatively, there are a number of drawbacks in its network training process and the resulting superpixels are not very regular or adaptive in size. 
Most recently, \cite{yang2020superpixel} uses fully convolutional network in an Encoder-Decoder structure to learn the association between pixels and superpixels on a regular image grid. As a result, it cannot produce the exact user-specified number of superpixels. Additionally, this method also requires a post-processing technique to remove any unwanted superpixels that are tiny in sizes.

%% differences next:
\smallskip
\textbf{Distinctions in our approach.} Our technique is close in philosophy to that of~\cite{tu2018learning} and~\cite{liu2011entropy}. Both our work and~\cite{tu2018learning} seek to learn pixel-wise affinities, however our proposal has several major advantages. 
Firstly, the work of~\cite{tu2018learning} is limited by its own design -- as it only allows the learning of pairwise pixel affinities for one direction at a time. Therefore, the network needs to be applied twice to obtain both horizontal and vertical pixel affinities, which results only in a 4-connected affinity map. By contrast, our work learns a richer 8-connected affinity map within one training process. 
Secondly, the loss design of \cite{tu2018learning} requires applying a superpixel segmentation method for each training epoch, which is computationally costly. Additionally, it requires incorporating external edge information, which does not necessarily generalise well on the training set. In comparison, our work efficiently enforces boundary adherence by learning directly from the network without the need for superpixel segmentation or incorporation of external edge information throughout training. 
Finally, we introduce an efficient graph-based technique called Hierarchical Entropy Rate Segmentation (HERS). It can produce highly adaptive superpixels instantaneously without the need for any parameter. Notably, we offer a computational complexity of $\mathcal{O}(|\nodes|)$, as opposed to $\mathcal{O}(|\nodes|^{2}\log (|\nodes|))$~\cite{liu2011entropy}. % Finally, our HERS avoid the non-differentiability issue in the training process as in ~\cite{jampani2018superpixel} and allows for using standard deep nets as in ~\cite{yang2020superpixel} but with better efficiency. %\Angie{we can slightly reduce this section if we need space}.

\section{The Proposed Methodology}
In this section, we present the two core parts of our proposed graph-based superpixel segmentation framework: i) our proposed \textbf{Deep Affinity Learning (DAL)} neural network architecture for obtaining deeply learned pixel affinities, and ii) our proposed \textbf{Hierarchical Entropy Rate Segmentation (HERS)} algorithm. An overview of our proposal is displayed in Figure~\ref{fig:network_flow}.

\subsection{Deep Affinity Learning}
In the first stage of our proposed framework, our goal is to design an effective and efficient network training scheme that produces an 8-connected affinity map $A\in\mathbb{R}^{8\times H\times W}$ for any input image of height $H$ and width $W$. The total number of pixels in an image is given by $N=W\times H$. For each pixel $\pixel_{i}$, we compute the affinities between $\pixel_{i}$ and a maximum of 8 surrounding pixels (horizontally and diagonally) that lie in its closest neighbourhood $\mathcal{N}_{i}$. 

\smallskip 
\textbf{Network design.} Our proposed Deep Affinity Learning network, \ournet for short, consists of two parts. 
In the first part, a $7\times 7$ convolutional kernel is used in the first layer to capture both the horizontal and vertical changes in an image. This is followed by 3 standard residual blocks (ResBlock)~\cite{he2016deep}, each of which has a kernel size of 3 and there are 8 input and output channels. Using 3 ResBlocks would expedite the decrease of the training loss without inducing too much additional computational burden and extra model parameters. Using 8 input and output channels allows us to obtain an intermediate 8-channel affinity map that captures pairwise pixel proximity towards all possible directions.

In the second part, we seek to enforce the preservation of fine details and scene structures. To achieve this, we integrate the HED network structure from~\cite{xie2015holistically}. 
% Explain what are the side outputs from each of these five continuous blocks.
That is, we incorporate five continuous blocks of convolutional layers with increasing receptive field sizes (64, 128, 256, 512, 512) to capture neighbourhood information at varying scales. 
% Explain how the side outputs are fused together 
Each of these five varying scale blocks produces a side output of 8 channels through an additional convolutional layer. Finally, a weighted fusion layer is used to automatically learn how to combine the side outputs from multiple scales. Concretely, the weighted fusion layer is a convolutional layer with 40 ($5\times 8$) input channels and 8 output channels to produce the final 8-channel affinity map~$A$.

\smallskip 
\textbf{Loss function.} 
% Explain how the ground truth segmentation map T is obtained 
Each image comes with a ground truth segmentation mask $M\in\mathbb{R}^{H\times W}$, where $m_{ij}\in\left\{1,\ldots, \nclass \right\}$ denotes the pixel-specific class label, and $\nclass$ is the total number of classes in the image. By comparing the label of each pixel $\pixel_{i}$ with the labels of its maximally 8 surrounding pixels, we can transform $M$ into an 8-channel binary segmentation map $T\in\left\{0,1 \right\}^{8\times H\times W}$, indicating whether two pairwise pixels belong to the same class label or not. 

As such, the training loss can be quantified separately for both boundary pixels and non-boundary pixels according to $T$. The most common way of measuring the loss between the learned affinity map $A$ and the ground truth map $T\in\left\{0,1\right\}^{8\times H\times W}$ is through the following Binary Cross Entropy~(BCE) loss:
\vspace{-.5cm}
\begin{equation}\label{bce_loss}
\begin{aligned}
\mathcal{L}_{BCE}(A,T)=
&-\frac{1}{8N}\sum_{i=1}^{N}\sum_{j\in\mathcal{N}_{i}}(1-t_{ij})\log(1-a_{ij})\\
&-\frac{1}{8N}\sum_{i=1}^{N}\sum_{j\in\mathcal{N}_{i}}t_{ij}\log(a_{ij}),
\end{aligned}
\end{equation}
where 
%$N$ is the total number of pixels, $\mathcal{N}_{i}$ is the neighbourhood of pixel $i$ that contains its surrounding 8 pixels, 
$t_{ij}\in\left\{0,1\right\}$ denotes the ground truth relationship between pixel $\pixel_i$ and $\pixel_j$, and $a_{ij}\in\left[0, 1\right]$ denotes the learned affinity between pixel $\pixel_i$ and $\pixel_j$.

It can be seen that~\eqref{bce_loss} encourages the learned affinities to be zeros for boundary pixels, and ones for non-boundary pixels. However, the ground truth segmentations are often given for object detection instead of superpixel segmentation. Therefore, there is potentially no supervision information available in the heterogeneous regions within an object. 
%This is motivated by the fact that one seeks to capture structure-rich parts instead of large uniform areas. 
Motivated by this, we propose the following modified version of~\eqref{bce_loss}:
\vspace{-.4cm}
\begin{equation}\label{our_loss}
\begin{aligned}
\mathcal{L}_{\ournet}(A, T)=
&-\frac{1}{8N}\sum_{i=1}^{N}\sum_{j\in\mathcal{N}_{i}}(1-t_{ij})\cdot\log(1-a_{ij})\\
&-\frac{1}{8N}\sum_{i=1}^{N}\sum_{j\in\mathcal{N}_{i}}t_{ij}\cdot|\gausssim_{ij}-a_{ij}|,
\end{aligned}
\end{equation}
where $\gausssim_{ij}$ denotes the pre-computed pairwise pixel affinity between $\pixel_{i}$ and $\pixel_{j}$. 
% formula for how the Gaussian similarity is calculated 
We use the Gaussian similarity to compute the pairwise pixel affinity $\gausssim_{ij}$ as: $\gausssim_{ij}=-\exp\left\{{d(\pixel_{i}, \pixel_{j})}/{2\sigma^{2}}\right\}$~\cite{boykov2001experimental},
in which $\sigma$ is the bandwidth parameter and $d(\pixel_{i}, \pixel_{j})$ denotes the distance between pixel $\pixel_i$ and $\pixel_j$. It can be computed as the $\ell_{2}$ distance between the RGB pixel features. As such, the loss in \eqref{our_loss} would encourage the learning of boundary information just as in the BCE loss, whilst learning additional local pixel affinity information for the non-boundary pixels guided by Gaussian similarity. Note that this is very different to the BCE loss, in which all non-boundary pairwise pixel affinities are treated as 1s. Since the majority of pixels in an image are non-boundary pixels, training the network using~\eqref{our_loss} enjoys a substantial advantage over the use of the BCE loss.

% other possible variants of the loss function 
%\textbf{Why is $\mathcal{L}_{DAL}$ effective?} There are a few other additional modifications that one can consider. For example, we can add a weighting scheme to balance the boundary and non-boundary loss terms as is done in~\cite{xie2015holistically}. Alternatively, \cite{tu2018learning} multiplies the original BCE loss with an over-segmentation error which implicitly handles the data balance issue. Although it has been shown to be effective in encouraging boundary adherence, it is at the cost of having to compute the superpixel segmentation thus the over-segmentation error at every epoch of the training process. 
%Yet another potential alternative is to replace the ground truth $t_{ij}$ in the non-boundary loss term of ~\eqref{bce_loss} with the hand-crafted affinity $w_{ij}$, which removes the advantages of the learning affinities generalisation. %We further support our design in the experimental setting through a detailed ablation study. 

\subsection{Hierarchical Entropy Rate Segmentation}
%% first describe how it works originally in the ERS paper 
After obtaining an 8-channel affinity map $A\in\mathbb{R}^{8\times H\times W}$ from our trained  network, the second part of our proposed framework is to use the extracted rich information to generate superpixels. To achieve this, we introduce Hierarchical Entropy Rate Segmentation (HERS) algorithm next.

The idea is to represent the extracted information as an undirected graph $\mathcal{G}=(\mathcal{V},\mathcal{E})$, where $\mathcal{V}$ denotes the set of nodes on the graph that correspond to the image pixels. The adjacent nodes / pixels are connected by a set of edges $\mathcal{E}$, whose weights reflect their pairwise similarities. We wish to select a subset of edges $E\subset \edges$ from the graph~$\graph=(\nodes, \edges)$ such that the entropy rate of the segmentation is maximised. It has been shown in~\cite{liu2011entropy} that higher entropy rate corresponds to more compact and homogeneous clusters. The entropy rate is used to describe the uncertainty of a stochastic process. By modelling a graph $\graph=(\nodes, \edges)$ as a first-order Markov process, we can obtain the entropy rate of the graph as a conditional entropy.

In this framework, the transition probability between a pair of nodes $v_{i}$ and $v_{j}$ can be expressed as $p_{ij}=w_{ij}/w_{i}$, in which $w_{i}=\sum_{k:e_{ik}\in \edges}w_{ik}$ denotes the sum of edge weights connecting to node $v_{i}$. Here $w_{ij}$ denotes the pairwise similarity between $v_i$ and $v_j$, which can be readily obtained from the affinity map $A$ as $(a_{ij}+a_{ji})/2$. A stationary distribution of the Markov process can be given as follows:
\begin{equation}
\vspace{-.4cm}
\bm{\mu}=\left[\mu_{1}, \mu_{2}, \ldots, \mu_{|\nodes|}\right]=\left[\frac{w_{1}}{w_{T}},\frac{w_{2}}{w_{T}},\ldots,\frac{w_{|\nodes|}}{w_{T}} \right]^{\mathsf{T}}.
\end{equation}
Then the entropy rate of the graph can be expressed as
\begin{equation}
\mathcal{H}(\edges)=-\sum_{i}\frac{w_{i}}{w_{T}}\sum_{j}\frac{w_{ij}}{w_{i}}\log\frac{w_{ij}}{w_{i}},
\label{er}
\vspace{-.25cm}
\end{equation}
where $w_{T}=\sum_{i=1}^{|V|}w_{i}$ denotes the sum of weights across all nodes.
Our goal is to sequentially select edges, guided by the entropy rate criterion in~\eqref{er}, in order to \textit{form balanced superpixels that are adaptive in sizes}. To this end, we propose to use Bor\r{u}vka's algorithm to optimise solely the entropy rate of the graph, whilst obtaining balanced and adaptive superpixels. 

Bor\r{u}vka's algorithm is one of the oldest algorithms designed to identify the minimum spanning tree (MST) of a graph $\graph=(\nodes, \edges)$. It simultaneously goes through the following steps for all currently formed trees in the graph: i) identify the best edge for each tree, which consists of one node or a set of nodes that are already connected by existing edges, and ii) add those identified edges to the forest until one minimum spanning tree is formed connecting all nodes in the graph. 
%% The advantages of Boruvka and how we can use it to solve the ERS problem 
Bor\r{u}vka's algorithm enjoys a number of desirable properties. Firstly, instead of having to sort all the edges globally as in the lazy greedy algorithm adopted by~\cite{liu2011entropy}, Bor\r{u}vka's algorithm finds the best edges for all trees simultaneously. As a result, we can offer a  computational complexity of $\mathcal{O}(|\nodes|)$, as opposed to $\mathcal{O}(|\nodes|^{2}\log (|\nodes|))$ as in~\cite{liu2011entropy}. In addition, the parallel edge selection process in Bor\r{u}vka's algorithm allows for the generation of very balanced trees. Finally, by recording the orders in which the edges are added, a hierarchical structure can be formed which allows any number of trees to be generated instantaneously.  
% An illustration of these advantages is displayed in Figure~\ref{fig::Borruvka}.
% \begin{figure}[htbp!]
% 	\centering
% 	\begin{subfigure}[b]{.32\linewidth}
% 		\includegraphics[width=\linewidth]{imgs/baby_exp1.png}
% 		\caption{}
% 	\end{subfigure}
% 	\begin{subfigure}[b]{.32\linewidth}
% 		\includegraphics[width=\linewidth]{imgs/baby_exp3.png}
% 		\caption{}
% 	\end{subfigure}
% 	\begin{subfigure}[b]{.32\linewidth}
% 		\includegraphics[width=\linewidth]{imgs/baby_exp4.png}
% 		\caption{}
% 	\end{subfigure}
% 	\caption{A simple example that illustrates the efficiency and parallelisability of Bor\r{u}vka's algorithm. It only needs one iteration to arrive at the final partitioning, whereas the lazy greedy algorithm in~\cite{liu2011entropy} requires four iterations.} \label{fig::Borruvka}
% \end{figure}

The idea of using Bor\r{u}vka's algorithm for the purpose of superpixel segmentation has been previously utilised in~\cite{wei2018superpixel}. However, \cite{wei2018superpixel} considers edge weights as dissimilarities between pairs of nodes, whereas we associate edge weights with their potential contributions to the overall entropy rate of the graph. Concretely, the edge weight can be interpreted in terms of its contribution to the overall entropy rate of the graph, if it were to be added to the current set of selected edges $E$. Using the learned affinities $a_{ij}\in A$ for the edge weight $w_{ij}$ between node $v_{i}$ and $v_{j}$, the contribution of each edge $e_{ij}$ to the overall entropy rate can be calculated as:
\begin{equation*}
\begin{aligned}
&\mathcal{H}(E\cup\left\{e_{ij}\right\})-\mathcal{H}(E)\\
=&(w_{ij}+\mu_{i})\log (w_{ij}+\mu_{i})+(w_{ij}+\mu_{j})\log (w_{ij}+\mu_{j})\\
&-\mu_{i}\log\mu_{i}-\mu_{j}\log\mu_{j}-2w_{ij}\log w_{ij}.	
\end{aligned}
\vspace{-.2cm}
\end{equation*}

We refer to our proposal of sequential edge selection guided by entropy rate via Bor\r{u}vka's algorithm as \textbf{Hierarchical Entropy Rate Segmentation (HERS)}. The algorithmic form of HERS is presented in Algorithm~\ref{algo_boruvka}. Note that the number of superpixels $K$ is not necessarily required as an input to the algorithm. Instead, a segmentation with any number of superpixels can be obtained instantaneously from the hierarchy that is constructed via Algorithm~\ref{algo_boruvka}.
\begin{algorithm}[t]
	\SetAlgoLined
	\LinesNumbered
	\SetKwInOut{Input}{Input}
	\Input{Graph: $\graph=(\nodes, \edges)$
	}
	\textbf{Output}: A minimum spanning tree $T$; A set of selected edges $E\subset \edges$\\
	
	\SetKwProg{Function}{function}{}{end}
	Constructed tree: $T=\emptyset$; 
	Selected edges: $E=\emptyset$;
	Unselected edges: $U=\edges$\\
	\SetKwRepeat{Do}{do}{while}
	\Function{Bor\r{u}vka$(\graph=(\nodes, \edges))$}{
		\Do{$|T|>1$}{
			- Identify the currently formed trees $T_{1}, T_{2}, \ldots, T_{k}$ where $k\leq |\nodes|$\\
			- Find the best outgoing edge for each tree 
			%$T_{i}$ ($i=1,2,\ldots, k$):
			$e^{\star}(T_{i})=\underset{e_{ij}\in U_{i}}{\arg\max} \text{ }\mathcal{H}(E\cup \left\{e_{ij}\right\}) - \mathcal{H}(E),$
			where $U_{i}$ denotes the unselected outgoing edges for tree $T_{i}$\\
			- Sort the best outgoing edges in descending order: $e^{*}(T_{1}), e^{*}(T_{2}), \ldots, e^{*}(T_{k})$ \\
			\For{$\ell=1,\ldots, k$}{
				Add the $\ell$-th outgoing edge to $E$;\\
				$K \leftarrow K-1$; 
			}
		}
	}
	\caption{HERS Superpixels}
	\label{algo_boruvka}
\end{algorithm}

%------------------------------------------------------------------------

\section{Experimental Results}\label{sec_experiments}
In this section, we describe in detail the range of experiments that we conducted to demonstrate the effectiveness and efficiency of our proposed methodology.
\begin{figure*}[ht!]
	\centering
	\begin{subfigure}[b]{.15\textwidth}
		%% original 
		%
		\includegraphics[width=\linewidth, height=.5\linewidth]{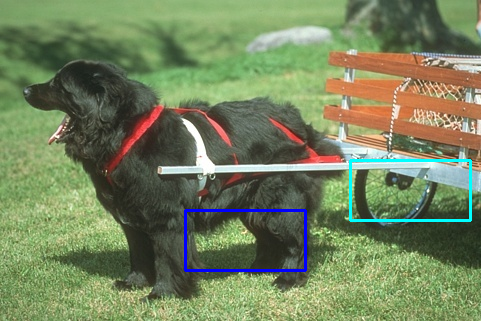}
		\includegraphics[width=.485\linewidth]{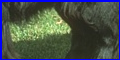}
		\includegraphics[width=.485\linewidth]{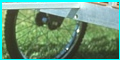}
		\includegraphics[width=\linewidth, height=.5\linewidth]{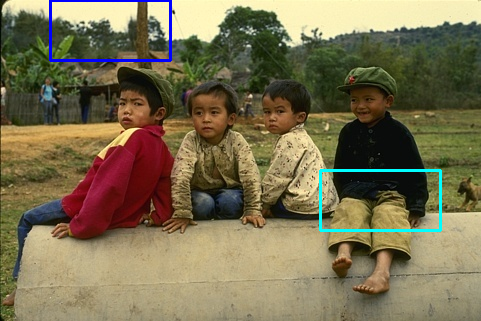}
		\includegraphics[width=.485\linewidth]{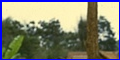}
		\includegraphics[width=.485\linewidth]{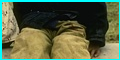}
		\includegraphics[width=\linewidth, height=.5\linewidth]{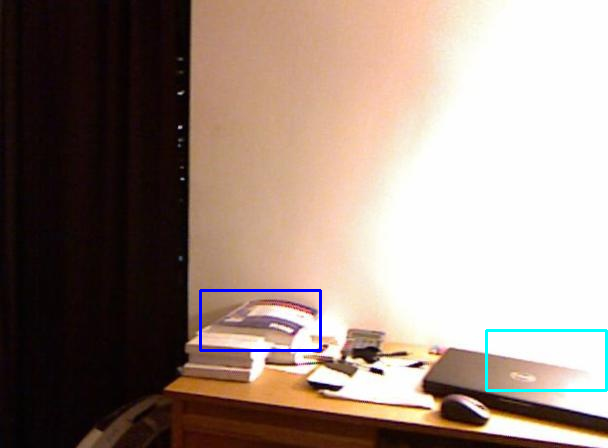}
		\includegraphics[width=.485\linewidth]{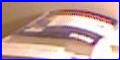}
		\includegraphics[width=.485\linewidth]{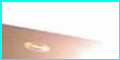}
		\caption{Original.}
	\end{subfigure}
	\begin{subfigure}[b]{.15\textwidth}
		%% ers
		%
		\includegraphics[width=\linewidth, height=.5\linewidth]{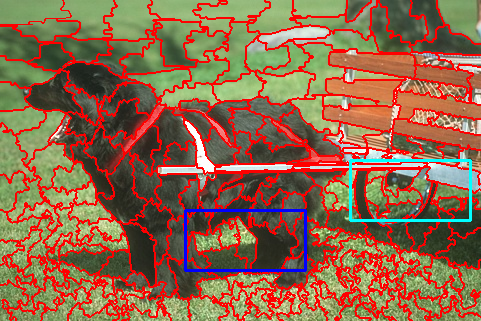}
		\includegraphics[width=.485\linewidth]{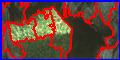}
		\includegraphics[width=.485\linewidth]{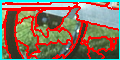}
		\includegraphics[width=\linewidth, height=.5\linewidth]{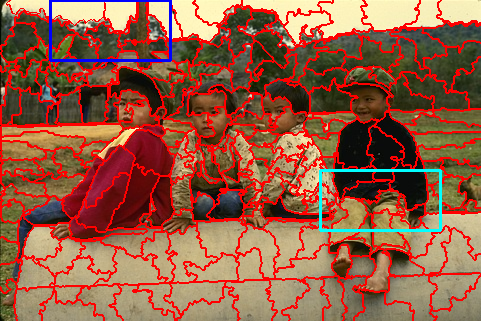}
		\includegraphics[width=.485\linewidth]{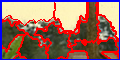}
		\includegraphics[width=.485\linewidth]{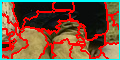}
		\includegraphics[width=\linewidth, height=.5\linewidth]{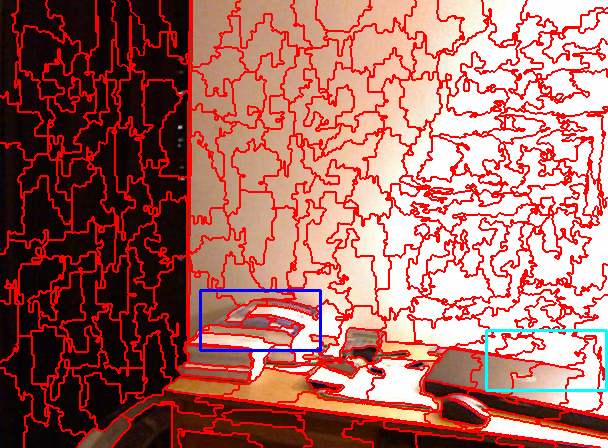}
		\includegraphics[width=.485\linewidth]{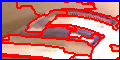}
		\includegraphics[width=.485\linewidth]{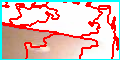}
		\caption{ERS.}
	\end{subfigure}
	\begin{subfigure}[b]{.15\textwidth}
		%% seal
		%
		\includegraphics[width=\linewidth, height=.5\linewidth]{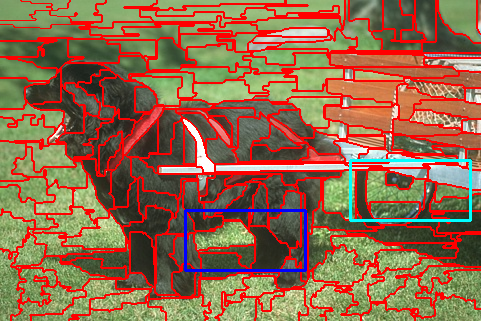}
		\includegraphics[width=.485\linewidth]{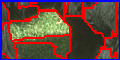}
		\includegraphics[width=.485\linewidth]{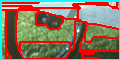}
		\includegraphics[width=\linewidth, height=.5\linewidth]{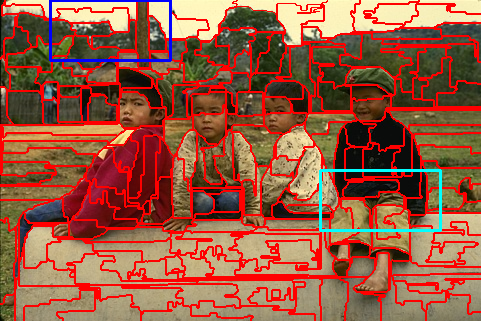}
		\includegraphics[width=.485\linewidth]{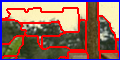}
		\includegraphics[width=.485\linewidth]{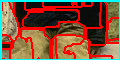}
		\includegraphics[width=\linewidth, height=.5\linewidth]{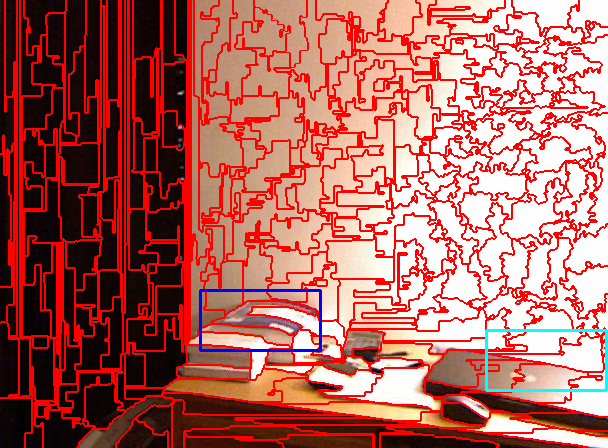}
		\includegraphics[width=.485\linewidth]{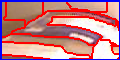}
		\includegraphics[width=.485\linewidth]{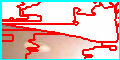}
		\caption{SEAL-ERS.}
	\end{subfigure}
	\begin{subfigure}[b]{.15\textwidth}
		%% SP-FCN 
		%
		\includegraphics[width=\linewidth, height=.5\linewidth]{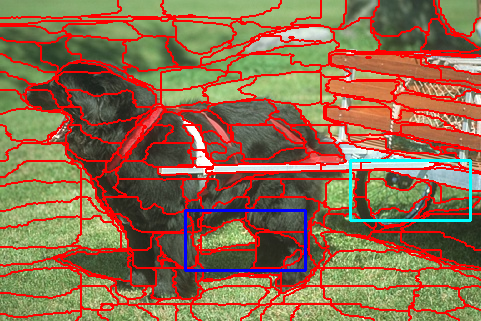}
		\includegraphics[width=.485\linewidth]{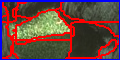}
		\includegraphics[width=.485\linewidth]{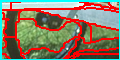}
		\includegraphics[width=\linewidth, height=.5\linewidth]{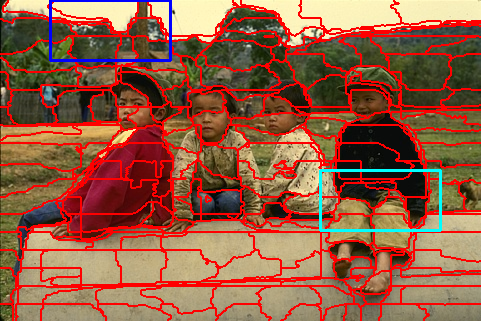}
		\includegraphics[width=.485\linewidth]{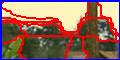}
		\includegraphics[width=.485\linewidth]{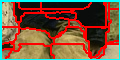}
		\includegraphics[width=\linewidth, height=.5\linewidth]{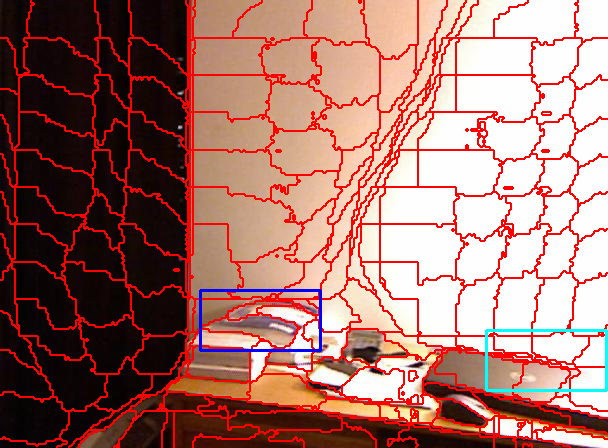}
		\includegraphics[width=.485\linewidth]{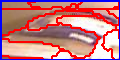}
		\includegraphics[width=.485\linewidth]{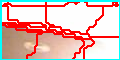}
		\caption{SP-FCN.}
	\end{subfigure}
	\begin{subfigure}[b]{.15\textwidth}
		%% SSN 
		%
		\includegraphics[width=\linewidth, height=.5\linewidth]{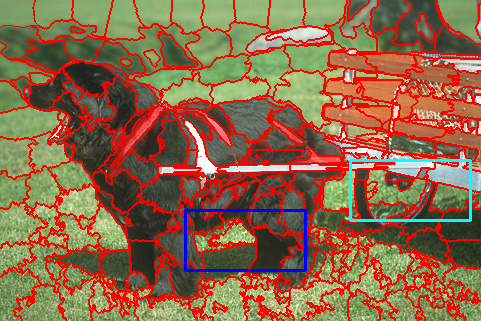}
		\includegraphics[width=.485\linewidth]{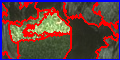}
		\includegraphics[width=.485\linewidth]{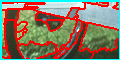}
		\includegraphics[width=\linewidth, height=.5\linewidth]{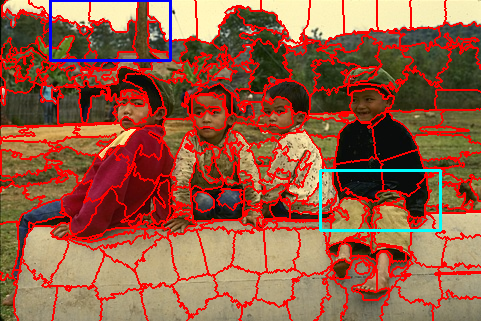}
		\includegraphics[width=.485\linewidth]{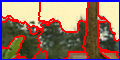}
		\includegraphics[width=.485\linewidth]{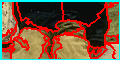}
		\includegraphics[width=\linewidth, height=.5\linewidth]{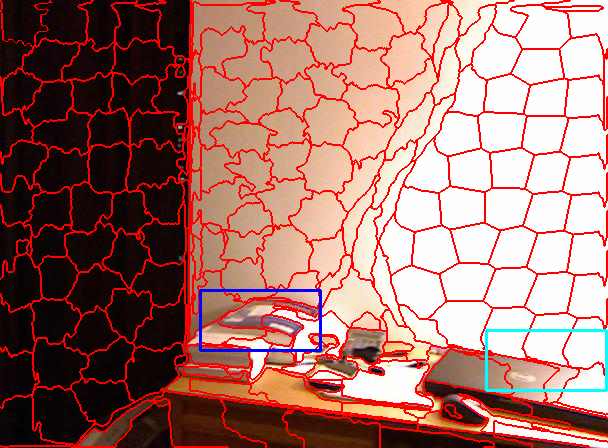}
		\includegraphics[width=.485\linewidth]{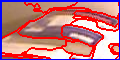}
		\includegraphics[width=.485\linewidth]{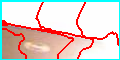}
		\caption{SSN.}
	\end{subfigure}
	\begin{subfigure}[b]{.15\textwidth}
		%% ours 
		%
		\includegraphics[width=\linewidth, height=.5\linewidth]{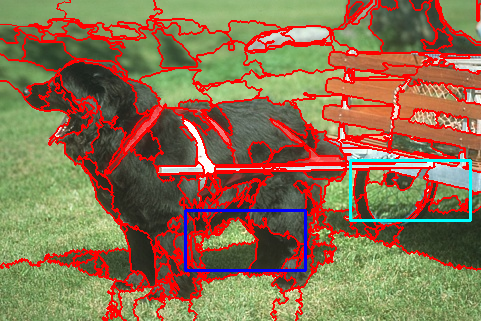}
		\includegraphics[width=.485\linewidth]{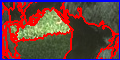}
		\includegraphics[width=.485\linewidth]{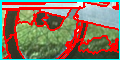}
		\includegraphics[width=\linewidth, height=.5\linewidth]{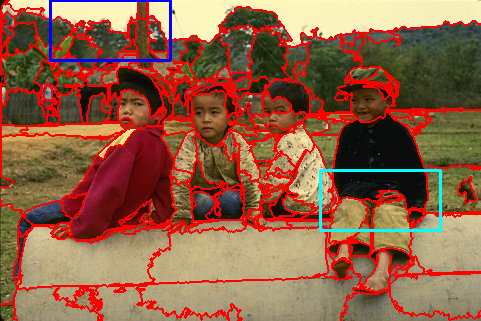}
		\includegraphics[width=.485\linewidth]{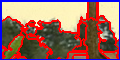}
		\includegraphics[width=.485\linewidth]{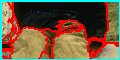}
		\includegraphics[width=\linewidth, height=.5\linewidth]{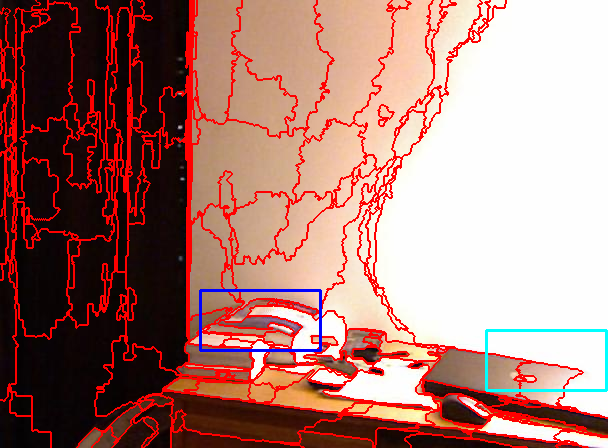}
		\includegraphics[width=.485\linewidth]{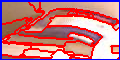}
		\includegraphics[width=.485\linewidth]{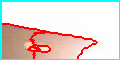}
		\caption{Ours.}
	\end{subfigure}
	\vspace{-0.2cm}
	\caption{Visual comparison of our technique against a number of state-of-the-art algorithms using 200 superpixels. The selected images show variations in scenes from BSDS500 and NYUv2 datasets.} % We highlight some regions in the zoom-in views, where our technique better captures the fine details and enjoys strong boundary adherence whilst avoiding partitioning homogeneous areas.} 
	\vspace{-0.4cm}
	\label{fig:vis_comp1}
\end{figure*}

\textbf{Dataset Description \& Evaluation Protocol.} 
We evaluate the performance of our technique using two benchmark datasets for superpixels. We use The Berkeley Segmentation Dataset 500 (BSDS500)~\cite{arbelaez2010contour}. This dataset contains 500 images with provided ground truth segmentations. It provides a wide variety of outdoor scenes with different complex structures. We also use the NYU Depth Dataset V2 (NYUv2)~\cite{silberman2012indoor}, which consists of 1449 images with provided ground truth segmentations. The NYUv2 dataset provides a range of different indoor scenes. 

We first justify the model design and support the advantage of our technique through a set of ablation studies. 
%The first one compares hand-crafted affinities our learned affinities, whilst the second one is related to further support of our model design. 
We then compare our technique with the following state-of-the-art techniques: i) classic techniques: ERS~\cite{liu2011entropy}, SH~\cite{wei2018superpixel}, SLIC~\cite{achanta2012slic}, SNIC~\cite{achanta2017superpixels},  SEEDS~\cite{van2012seeds}, ETPS~\cite{yao2015real}; and ii) deep learning techniques:  SSN~\cite{jampani2018superpixel}, SEAL-ERS~\cite{tu2018learning} and SP-FCN~\cite{yang2020superpixel}. 
The most commonly used performance measures for evaluating superpixel segmentation algorithms include: Under-segmentation Error (UE)~\cite{van2012seeds}, Achievable Segmentation Accuracy (ASA)~\cite{liu2011entropy} and Boundary Recall (BR)~\cite{martin2004learning}. In particular, UE can be directly obtained from ASA as 1-ASA. Thus, we report the ASA and BR measures in our experiments. Additionally, we also report the Explained Variation (EV)~\cite{moore2008superpixel} that quantifies the variance within an image that is captured by the superpixels without relying on any ground truth labels. Explicit definitions of the metrics can be found in the supplementary material.

\textbf{Implementation \& Training Details.} 
%The implementation of our proposed methodology consists of two stages. %\footnote{Our implementation is available at: \url{https://github.com/anonymous}}. 
We learn an 8-channel affinity map by training our proposed \ournet network on the BSDS500 training set. We implement the network in PyTorch using the Adam~\cite{kingma2014adam} optimiser with $\beta_{1}=0.9$ and $\beta_{2}=0.999$. The \ournet network is trained for 5k epochs, in which the input images are cropped to have size 200 by 200. The initial learning rate is set to $1e-4$, and is reduced by a factor of 10 after 3k epochs. For all the competing methods, we use the code provided by the corresponding authors.

\subsection{Ablation Study}~\label{sec:ablation}
\begin{figure*}[ht!]
	%	\begin{center}
	%		\includegraphics[width=.33\textwidth, height=.25\textwidth]{imgs/perf_curves/bsds_ablation_perf_comp_ev_20May.pdf}
	%		\includegraphics[width=.33\textwidth, height=.25\textwidth]{imgs/perf_curves/bsds_ablation_perf_comp_br_20May.pdf}
	%		\includegraphics[width=.33\textwidth, height=.25\textwidth]{imgs/perf_curves/bsds_ablation_perf_comp_asa_20May.pdf}
	%	\end{center}
	\begin{center}
		\includegraphics[width=.33\textwidth, height=.24\linewidth]{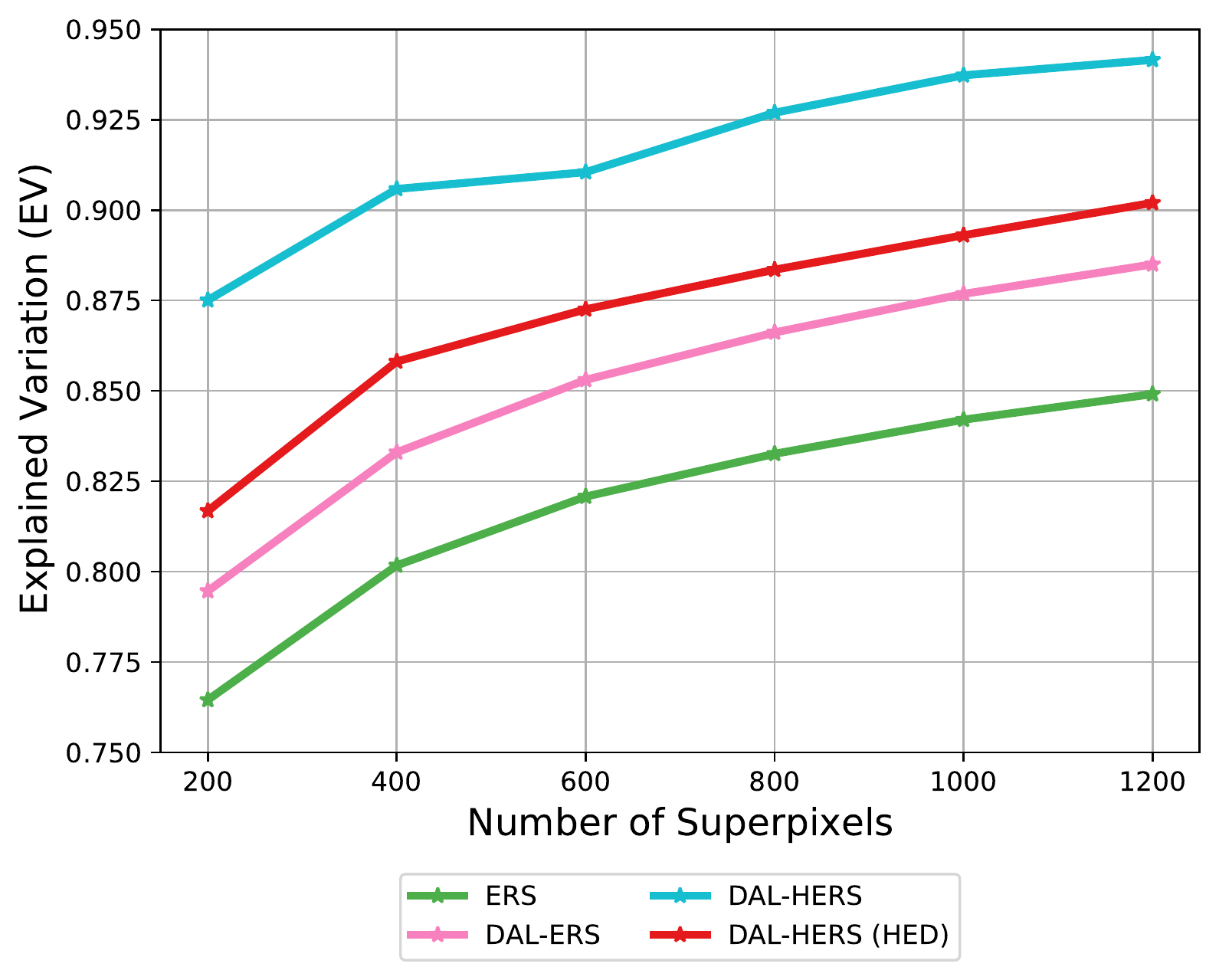}
		\includegraphics[width=.33\textwidth, height=.24\linewidth]{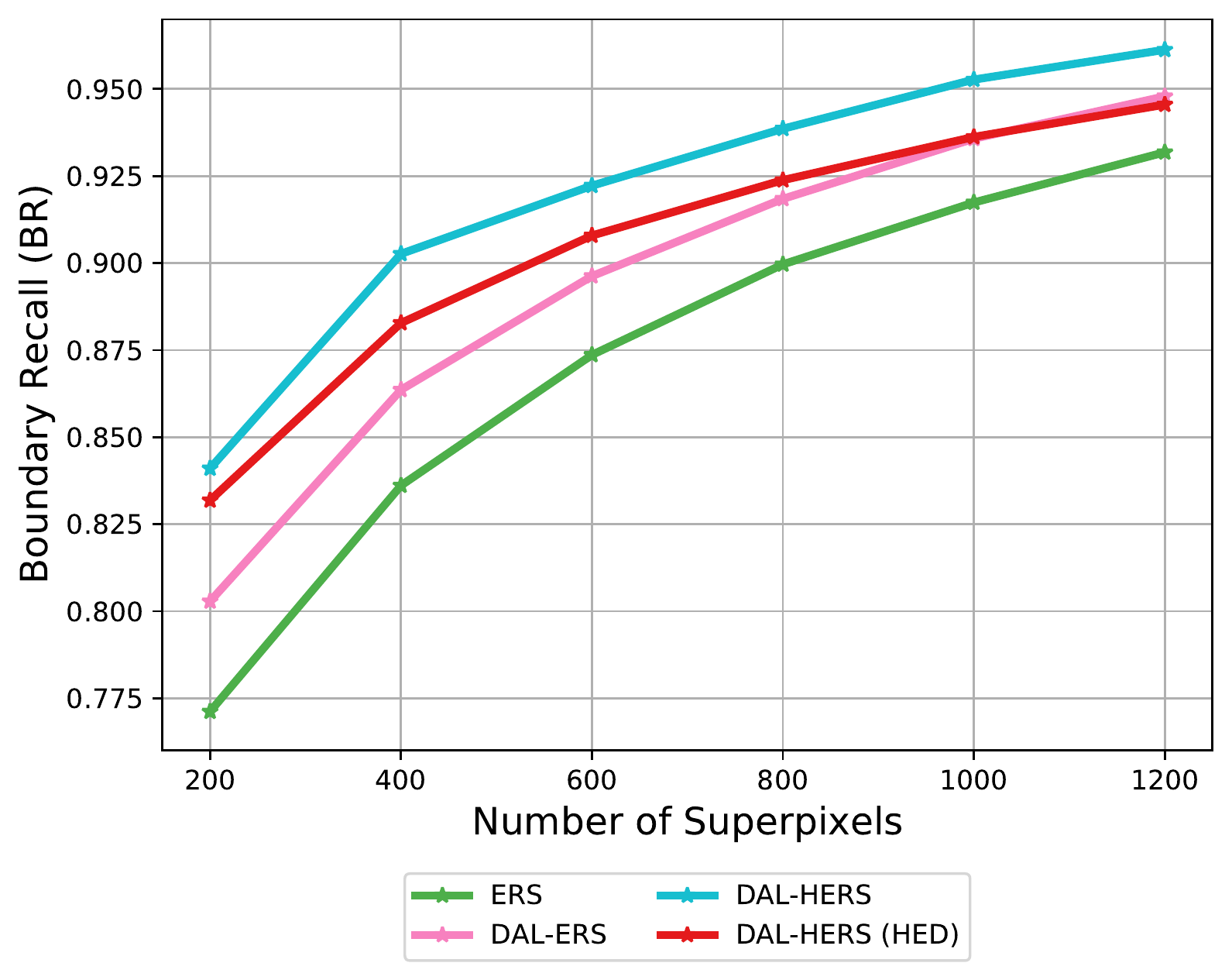}
		\includegraphics[width=.33\textwidth, height=.245\linewidth]{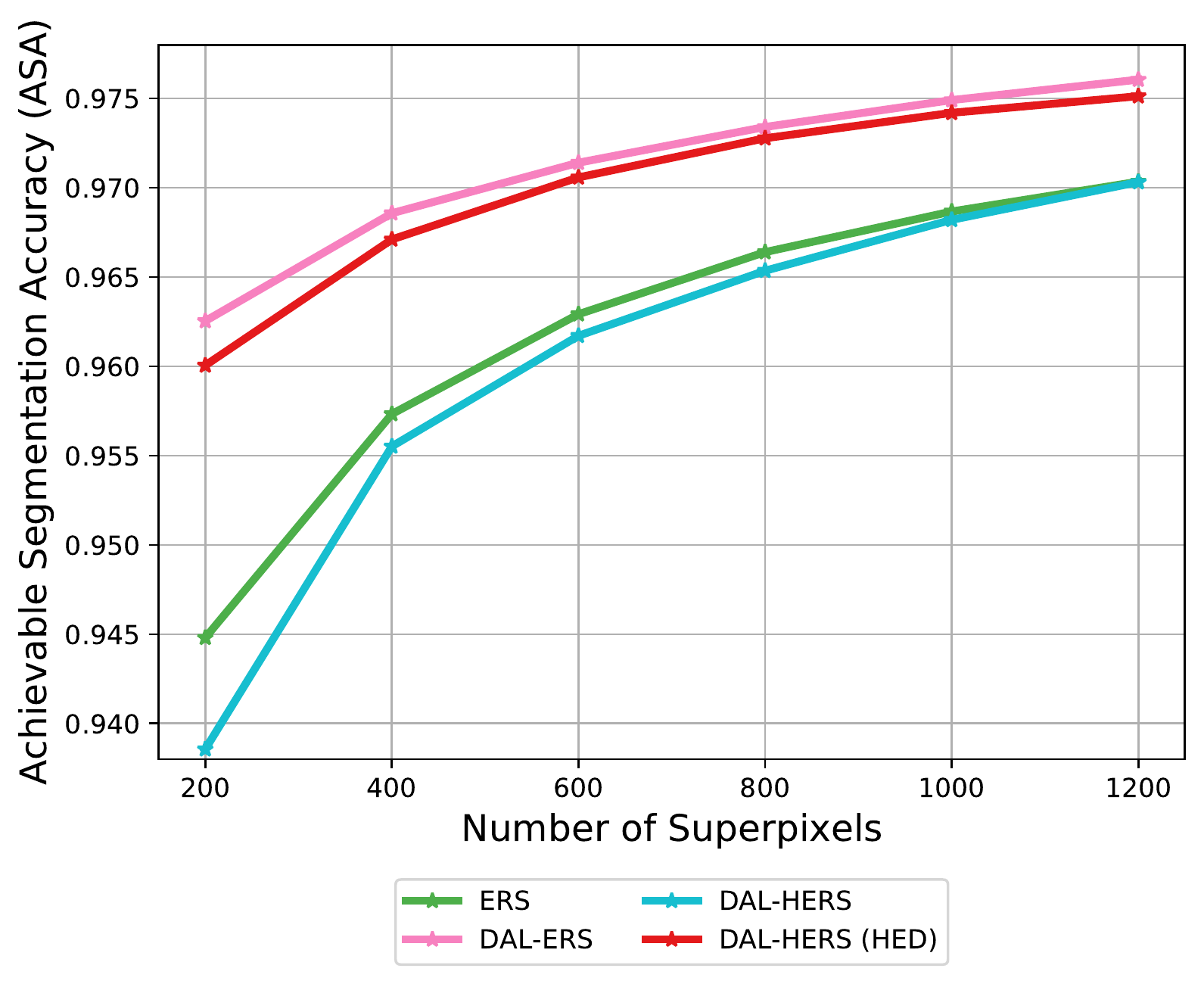}
	\end{center}
	\vspace{-.55cm}
	\caption{\textbf{Ablation study on the BSDS500 test set.} We compare the components in our proposed framework, which includes both the learned affinities from the DAL network and the HERS algorithm.}
	\label{fig_ablation}
	% The code that produces these two plots are in ./plotting_ablation.py
\end{figure*}

Our proposed methodology contains two main building blocks including the deeply learned affinities from the \ournet network and the proposed HERS algorithm. 
In this section, we evaluate the contribution from each of these components. The comparison across different variants are reported in Figure~\ref{fig_ablation} in terms of EV, BR, and ASA scores.

%In our proposed methodology, inspirations are drawn from the SEAL network structure in the affinity learning stage, and from the ERS and SH algorithm in the segmentation stage. Therefore, we use these methods as baselines in our ablation study to compare against several variants of our proposed methodology. In Figure~\ref{fig_ablation}, we present the performance comparison in terms of EV, BR, and ASA scores.

% Focus 1
\textbf{Benefit of the proposed DAL network.} The benefit of using learned affinities from the DAL network as opposed to handcrafted ones can be observed by comparing the performance of ERS (green) with that of DAL-ERS (pink) in Figure~\ref{fig_ablation}. ERS computes the pixel-wise affinities using the Gaussian kernel in which the RGB pixel values are used as features. Whereas DAL-ERS directly uses the learned affinities from the \ournet network as input to the ERS algorithm. It is clear that DAL-ERS (pink) largely outperforms the baseline ERS algorithm (green) in all three measures. % Note that the only difference in these three methods lies in how the affinities are learned / computed. 

\textbf{Benefit of the proposed HERS algorithm.} Next, we inspect the contribution of our proposed superpixel segmentation algorithm HERS. 
%which conducts hierarchical superpixel segmentation using both pixel affinities and pixel features. 
It is worth pointing out that HERS does not require any balancing term in the objective function and is therefore parameter-free. Whereas the balancing term in ERS plays a crucial role in avoiding extremely unbalanced superpixels.  

% DAL-HERS vs DAL-ERS
By comparing DAL-HERS (cyan) with DAL-ERS (pink), we can see that the former enjoys much higher EV and BR scores at the cost of a lower ASA score. 
% Explain this phenomenon
This trade-off can be explained by the fact that the superpixels produced by HERS are highly adaptive, i.e.\ \emph{it keeps large homogeneous regions of an image intact whilst over-segmenting the texture-rich regions}. Adaptive superpixels are arguably more desirable than superpixels whose sizes are agnostic to the semantics of the image. However such adaptive behaviour has certain implications on the performance measures. Since the superpixels produced by HERS adhere strongly to the object boundaries and preserves the homogeneous regions, it is to be expected that it enjoys very good EV and BR scores. However, having adaptive superpixels also means that a small ``leakage'' in the boundary adherence would incur a big under-segmentation error, which negatively correlates with the ASA score.

% DAL-ERS-Boruvka vs SH
%Since SH also uses the Bor\r{u}vka algorithm for superpixel segmentation whilst optimising for edge weights, we also consider it as a baseline and compare its performance to that of DAL-HERS. It is clear to see that DAL-HERS outperforms SH in both EV and BR scores whilst maintaining the same level of ASA scores. 

%Using handcrafted Gaussian affinities as input, we first compare ERS-Bor\r{u}vka (in pink) with SH-Bo\r{u}vka (in orange) and ERS (in blue). 
%% Compare ERS-Boruvka and SH-Boruvka
%There are two key differences that distinguishes our proposed ERS-Bor\r{u}vka from SH-Bor\r{u}vka. Firstly, instead of simply optimising the sum of edge weights, we use the entropy rate as objective which inherently encourages more compact superpixels. Secondly, both the pairwise affinities and the pixel features play important roles in ERS-Bor\r{u}vka.   
%As a result, ERS-Bor\r{u}vka enjoys better BR and EV score whilst mainting the same level of ASA as SH-Bor\r{u}vka.

\textbf{Benefit of external edge information.} 
%It is clear to see that there exists a trade-off between EV, BR, and ASA scores.
For the purpose of achieving a higher ASA score, we could compromise a small amount of the performance gain in terms of EV and BR scores. The ASA score can be improved by strengthening the adherence to object boundaries given by the ground truth segmentations. Therefore, we additionally divide the edge weight by HED edge probabilities, which corresponds to DAL-HERS (HED) in Figure~\ref{fig_ablation}. 
%In order to achieve this, it is important to note that the calculation of the ASA score is closely linked to the idea of superpixel ``leakage''. That is, even if most of the boundary pixels of a superpixel are identified correctly, a small amount of unidentified boundary pixels can potentially result in a big ``leakage'' area. 
As a result, we observe a notable increase in the ASA score of DAL-HERS (HED) as compared to that of DAL-HERS. The gain in the ASA score comes at a small cost of the EV and BR scores. This is again to be expected, because highlighting more boundary pixels would inevitably increase the false positives of boundary pixels thus decrease the BR and EV scores.
%In the rest of the paper, we will refer to DAL-ERS-Bor\r{u}vka and DAL-ERS-Bor\r{u}vka (HED) as \textbf{ERS-Bor\r{u}vka} and \textbf{ERS-Bor\r{u}vka (HED)} for simplicity.

% In this section, we compare our proposed method to various state-of-the-art classical and deep learning methods. 
%To make fair comparisons, we do not apply connectivity enforcement to the SP-FCN model and simply use the superpixels obtained from the affinity map generated using the pretrained model that the authors provide. This ensures fair comparison both in terms of network and loss design, and in terms of the number of generated superpixels are exactly the same as in other methods. 
%We include two versions of our proposal, \textbf{DAL-HERS} which takes the learned affinity map and the RGB pixel features as input and \textbf{DAL-HERS (HED)} which uses both the previous inputs and the HED edge map. For short, we refer to DAL-HERS as \emph{Ours} from now on. 

\begin{figure*}[t!]
	\centering
	\begin{subfigure}[b]{\textwidth}
		\centering
		\includegraphics[width=.33\textwidth, height=.24\linewidth]{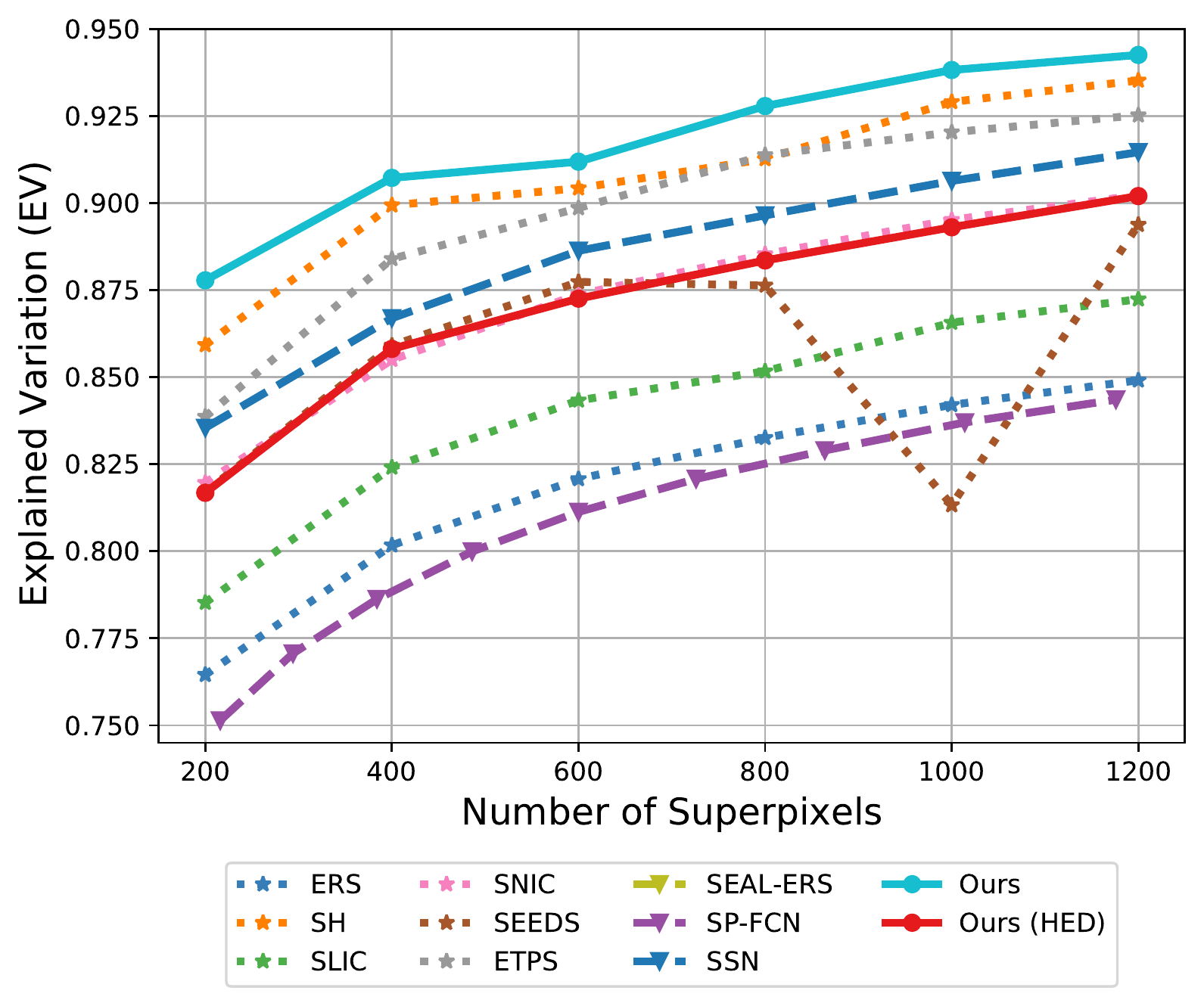}
		\includegraphics[width=.33\textwidth, height=.24\linewidth]{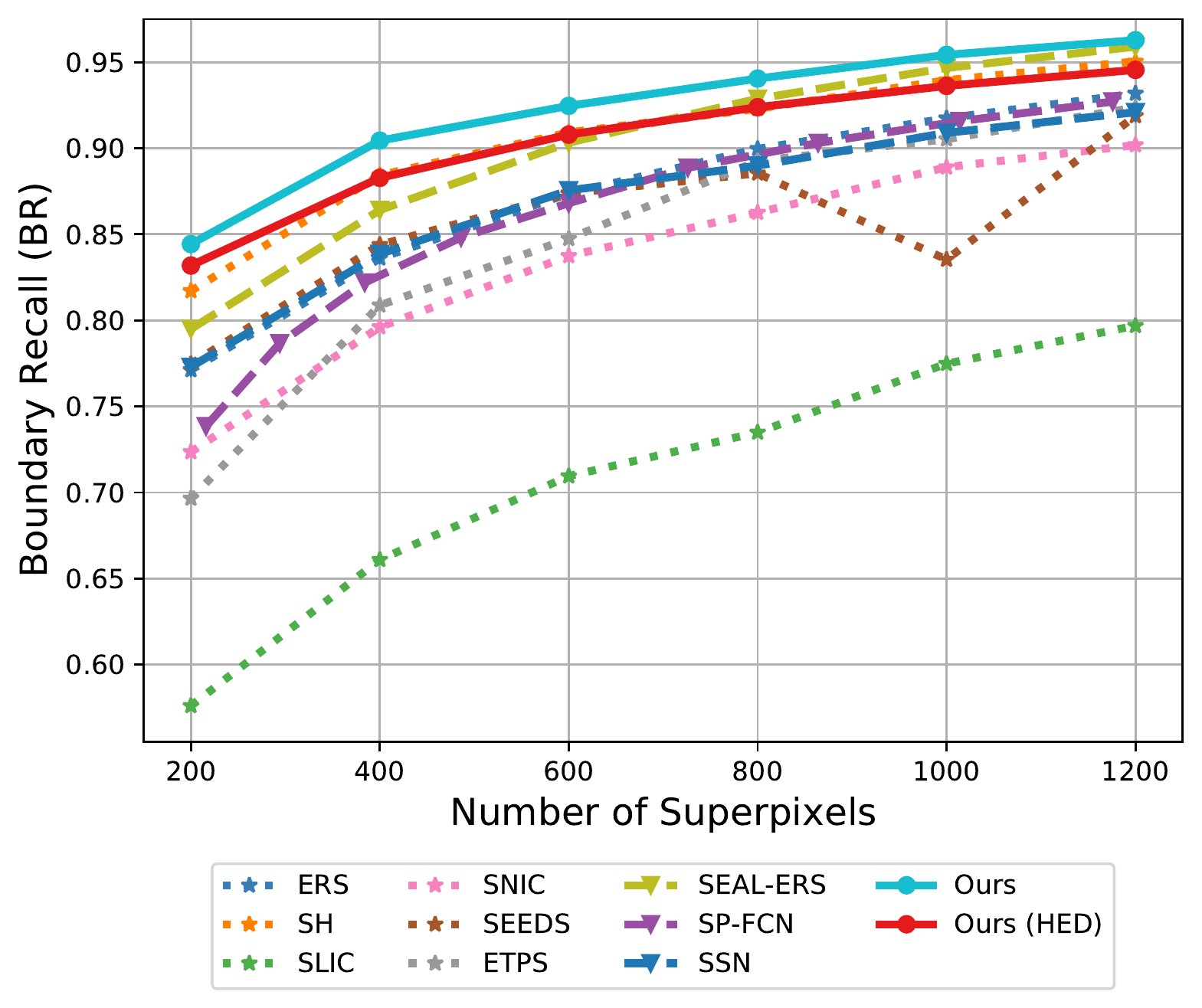}
		\includegraphics[width=.33\textwidth, height=.245\linewidth]{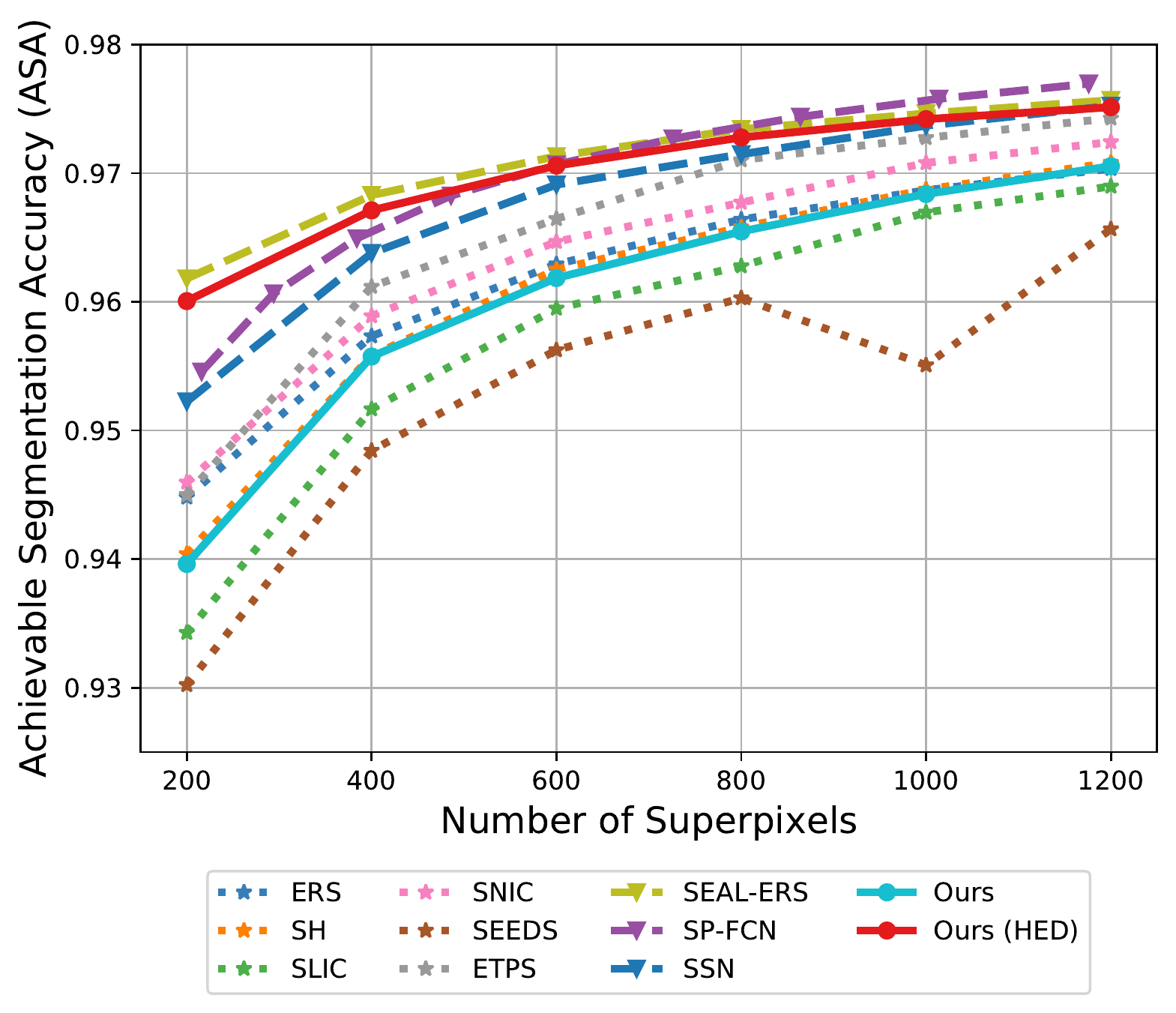}
		\caption{BSDS500.}
		\label{fig:y equals x}
	\end{subfigure}
	\hfill
	\begin{subfigure}[b]{\textwidth}
		\centering
		\includegraphics[width=.33\textwidth, height=.24\linewidth]{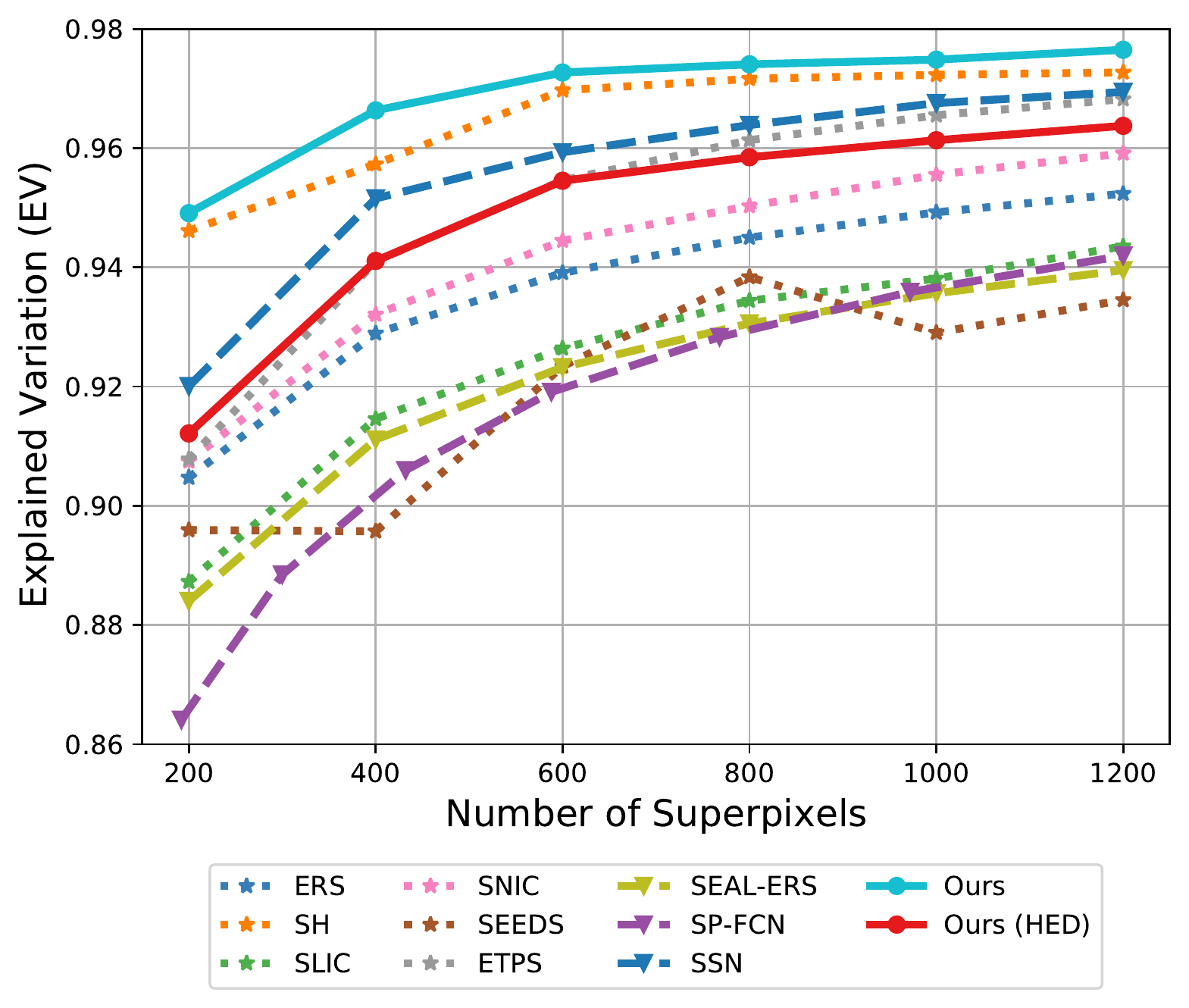}
		\includegraphics[width=.33\textwidth, height=.24\linewidth]{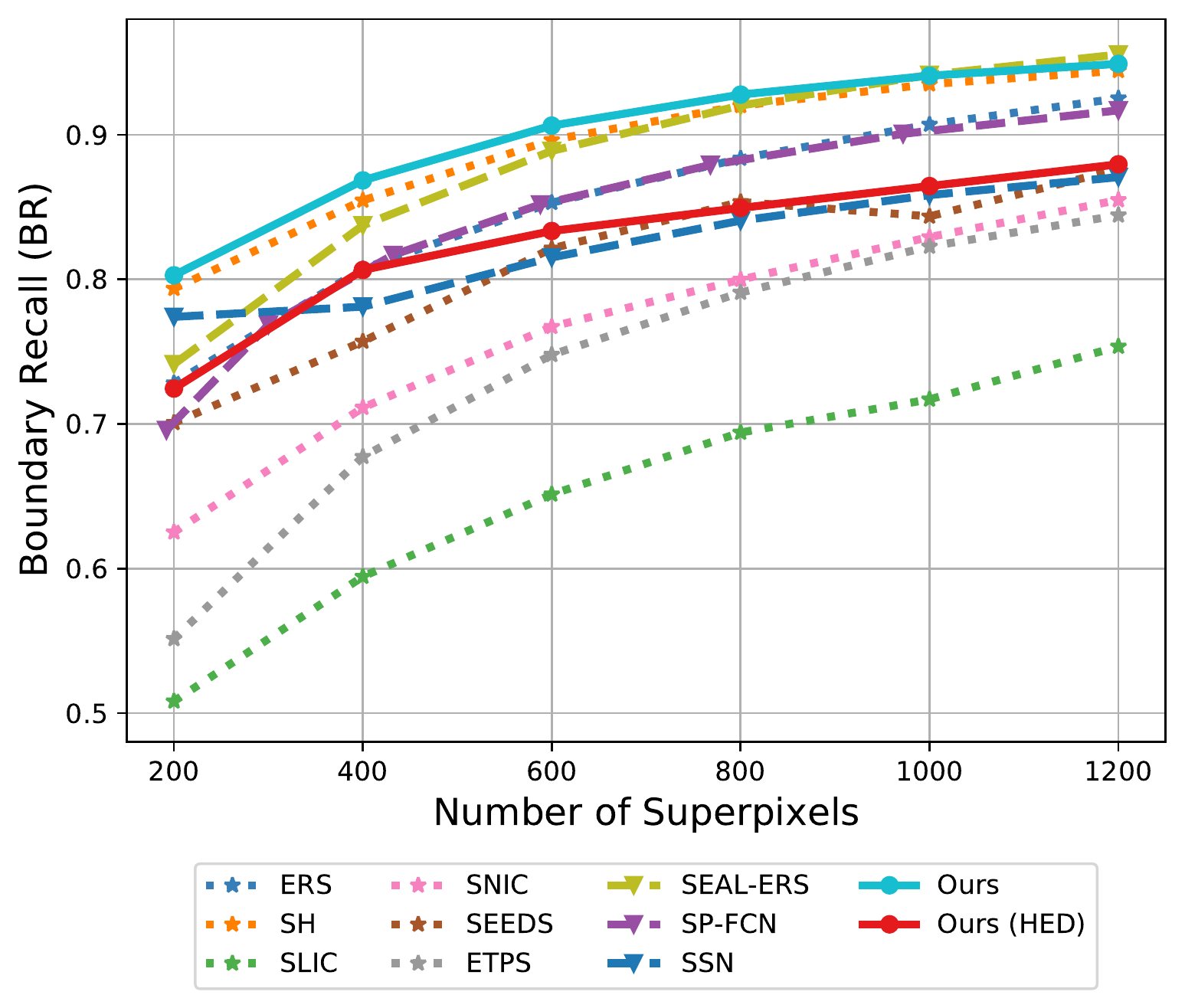}
		\includegraphics[width=.33\textwidth, height=.245\linewidth]{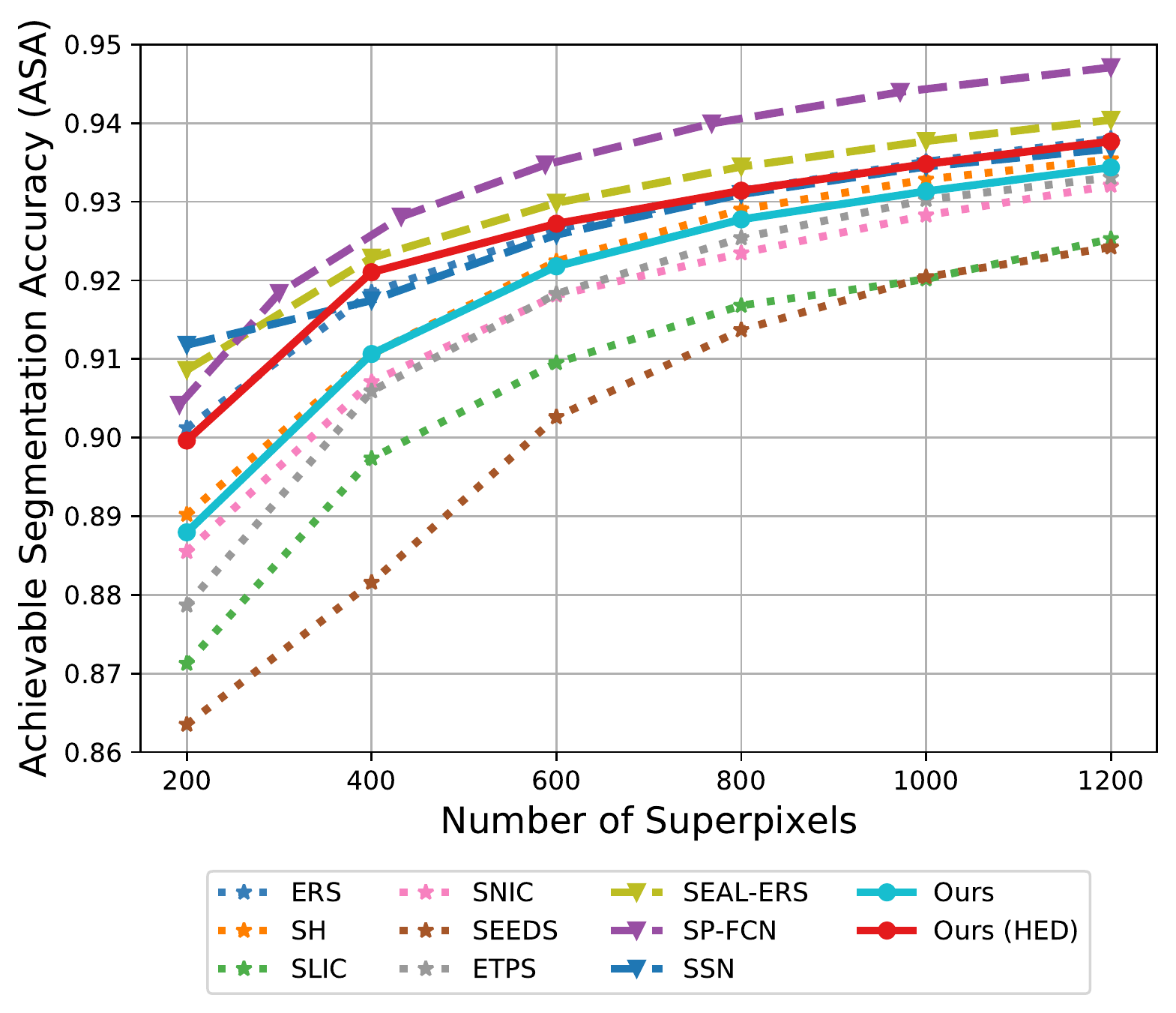}
		\caption{NYUv2.}
		\label{fig:three sin x}
		\vspace{-.35cm}
	\end{subfigure}
	\caption{Comparison of our approach (solid lines) against the state-of-the-art of classical approaches (dotted lines) and deep learning based approaches (dashed lines) for superpixels. The performance of various methods are evaluated using EV, BR and ASA scores over different numbers of superpixels.}
	\label{fig:sota_comparison}
	%\vspace{-.5cm}
\end{figure*}

\textbf{Quantitative performance.} Figure~\ref{fig:sota_comparison} shows the quantitative comparison across all methods on both the BSDS500 and the NYUv2 test set. We include both DAL-HERS and DAL-HERS (HED) in the comparison, and refer to them as Ours and Ours (HED) for short. 
%The experimental results for ERS, SEEDS, and SLIC are obtained using the code provided by~\cite{stutz2018superpixels}\footnote{https://github.com/davidstutz/superpixel-benchmark}. The results for other methods are obtained using the code that have been made publicly available by the corresponding authors.  
It can be seen that Ours outperforms all other competing methods in terms of EV and BR scores on both datasets. This means that the superpixels generated by our method fully captures the semantically homogenous regions of images, and at the same time strongly adheres to the object boundaries (see Figure~\ref{fig:vis_comp1}). Due to the highly adaptive nature of our superpixels, it does not lead to a high ASA score by design. 
The ASA score measures the overlap between the computed superpixels with the ground truth, where the ground truth labels are provided for object detection or semantic segmentation. As such, they do not delineate the fine details that our method captures so well, which results in the relatively low ASA score. To mitigate this, we could further incorporate HED edge information (as is explained in the ablation study).
%The excellent performance of these two measures comes at a cost of the ASA score, which can be remedied by incorporating the HED edge information as is detailed in the ablation study. 
As a result, Ours (HED) achieves competitive performance against the majority of the superpixel methods that are being compared to.

%However, the excellent performance in these two measures comes at a cost with slightly lower ASA scores. As has been shown previously in the ablation study in Section~\ref{sec:ablation}, the ASA scores can be improved by incorporating the HED edge information which results in a slight compromise from the EV and BR scores. 

\begin{figure}[t!]
	% Code for producing the following figure is in:
	% /home/hankui/Dropbox/Research/UnderReview/WACV-2022/Code/DGSS/plot_runtimes.py
	\centering
	\includegraphics[width=.85\linewidth, height=.55\linewidth]{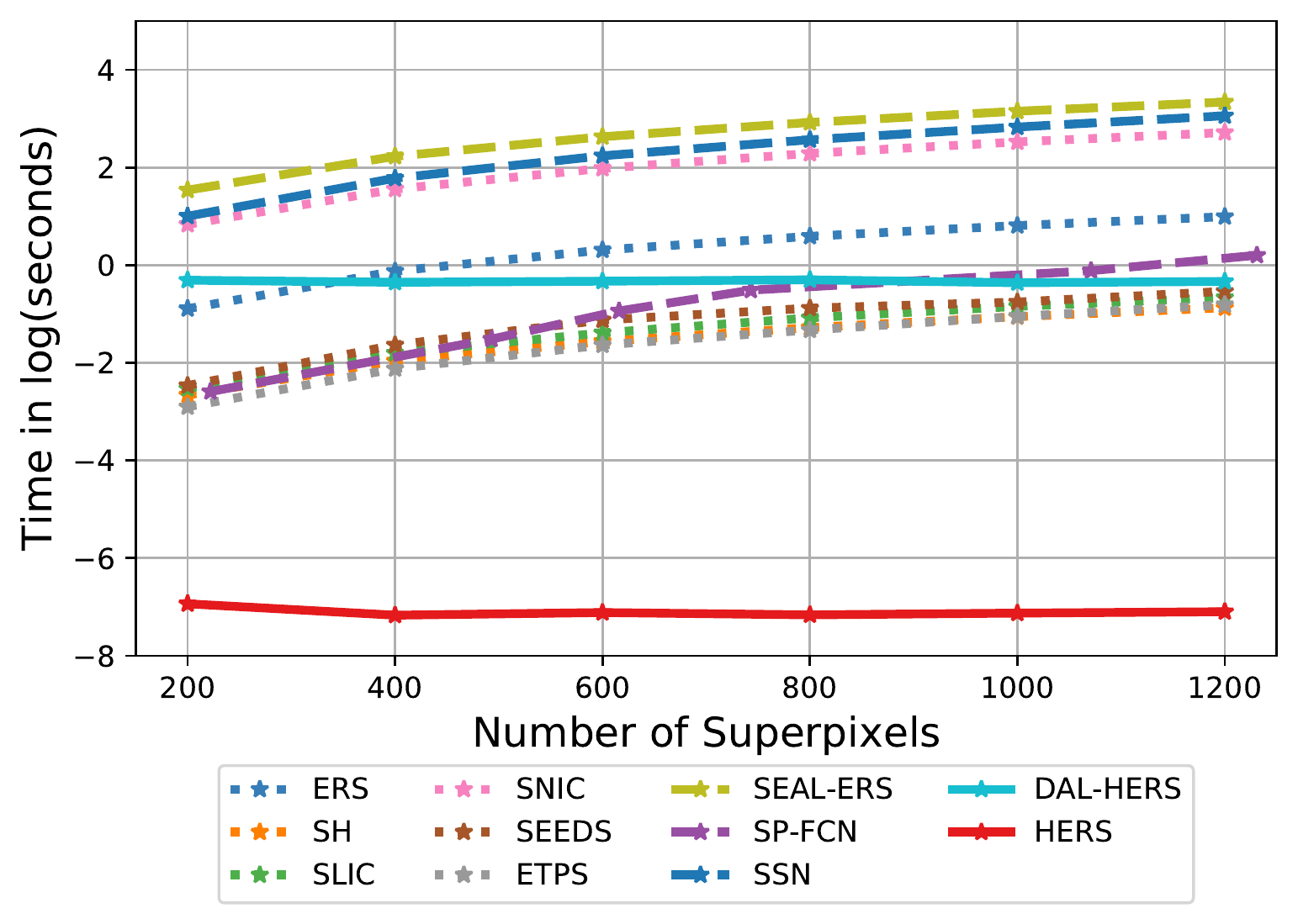}
	\vspace{-.4cm}
	\caption{Cumulative runtime in $\log_{e}(\text{seconds})$ on the whole BSDS500 test dataset.} 
	\vspace{-0.5cm}
	\label{fig_runtimes}
	\vspace{-.2cm}
\end{figure}

\textbf{Qualitative performance.} In addition to the quantitative advantages exhibited by our proposed schemes, additional advantages can be demonstrated via visual comparisons of the segmentation results. 
%In Figure~\ref{fig:vis_comp1}, we illustrate the difference of various methods by highlighting the segmented boundaries. 
In Figure~\ref{fig:vis_comp1}, we observe several advantages of our method over the compared ones. 
Firstly, our technique prevents the over-segmentation of homogeneous regions in the scene. Clear examples of this effect can be seen in the grass on the first row, the sky on the second row, and the walls on the third and fourth rows. 
In all images, our method preserves fine details on the objects by focusing on rich-structure parts rather than uniform regions. 
Secondly, our technique displays the best boundary adherence. That is, our technique is able to better capture the object structures well. Examples of this property can be found by inspecting the zoomed-in views of the book and chair on the third and forth rows. We provide further visual comparisons in the supplementary material.

%Across all images in Figure~\ref{fig:vis_comp1}, we note that the highly homogeneous regions (e.g.\ sky, clothes, light, window) of an image are left intact by our method, whereas all other methods segment and produce superpixels of roughly the same size across an image. At the same time, our method over-segments heterogeneous regions of an image where object boundaries lie in order to preserve the fine details. 

%It is worth pointing out that our method is the only method that produces superpixels that are adaptive in size and adhere well to the ground truth object boundaries.   

\subsection{Comparison to State-of-the-Art Techniques}

%\textcolor{blue}{-------------------------------------}

\textbf{Computational efficiency.} Another highly desirable property of any superpixel technique, as a stand-alone pre-processing tool, is the computational efficiency. We report the runtime (in log seconds) of our proposed DAL-HERS and HERS against other state-of-the-art methods in Figure~\ref{fig_runtimes}. To highlight the advantage of HERS, we report the cumulative runtime of each method for various numbers of superpixels. For example, the time comparison at 400 on the $x$-axis reports the time that a method requires to produce both 200 and 400 superpixels.

It is notable that the runtimes of both DAL-HERS and HERS are constant across various numbers of superpixels. This is because HERS constructs a hierarchical tree structure from which any number of superpixels can be extracted instantaneously, whereas all other methods have to run several times for multiple numbers of superpixels. 
When comparing HERS to other classical methods, it can be seen that HERS has a noticeable computational advantage against other methods. When comparing DAL-HERS to other deep methods, it is clear that it still enjoys competitive performance.
A main strength of our technique is the \textit{constant time} required to produce \textit{any number of superpixels especially as the numbers of superpixels increase.}

\begin{figure}[!t]
\centering
\includegraphics[width=0.49\textwidth, height=.22\textwidth]{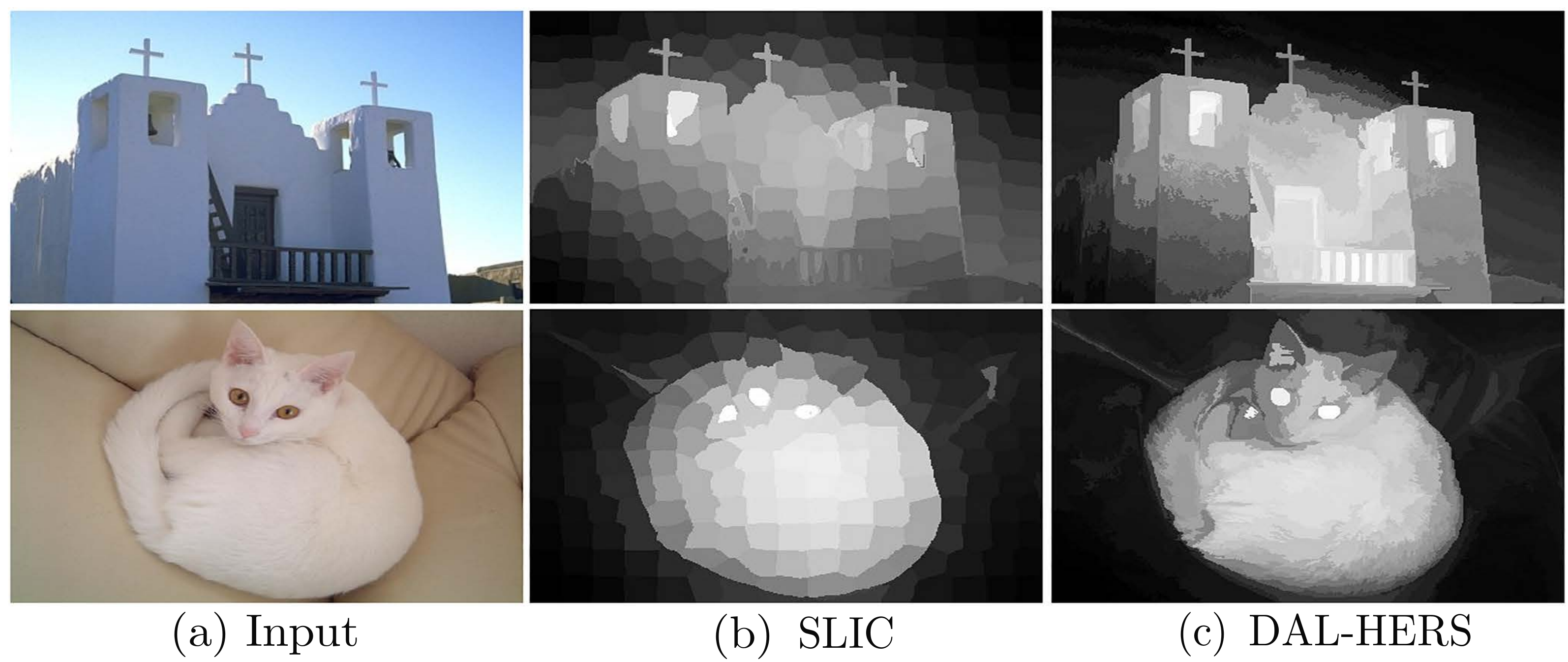}
\vspace{-0.5cm}
\caption{Visual comparison of saliency detection using SLIC and DAL-HERS with 200 superpixels on selected images from the ECSSD dataset~\cite{shi2015hierarchical}.} 
\vspace{-0.75cm}
\label{fig::saliency}
\end{figure}

\smallskip 
\textbf{Superpixels for saliency detection.}
To further support the advantage of DAL-HERS, we report results for the downstream task of saliency detection  for a  selection of images from the ECSSD dataset~\cite{shi2015hierarchical}. %For this experiment, 
we used the Saliency Optimisation (SO)~\cite{zhu2014saliency} technique as backbone, which originally uses SLIC to preprocess an image~\cite{achanta2012slic}. We replace SLIC with DAL-HERS and  present a visual comparison in Figure~\ref{fig::saliency}. One can see that DAL-HERS produces smoother outputs whilst preserving better edges than SLIC. 
%Metric-wise, we get \% improvement in terms of [metrics]. 
We provide a detailed explanation with metric and visual comparison in the supplementary material.

%------------------------------------------------------------------------

\vspace{-.2cm}
\section{Conclusion \& Future Work}
%% Conclusions 
In this paper, we present a graph-based framework consisting of a network for obtaining deeply learned affinities, and an efficient superpixel segmentation method for producing adaptive superpixels. Through experimental results, we show that our technique compares favourably against state-of-the-art methods. Moreover, unlike existing methods demanding linear time with respect to various numbers of user-specified superpixels, our technique exhibits constant time which makes it appealing to be used as a preprocessing tool.
%% Future work 
For future  work, our method could be tailored to various computer vision tasks such as semantic segmentation, saliency detection, and stereo matching. In addition, it would be interesting to combine our deeply learned affinities with other graph-based superpixel methods. 

%Systematic investigation in different deeply learned affinity maps (Ours, SEAL, SP-FCN)
\vspace{-.2cm}

\section*{Acknowledgements}
HP acknowledges the financial support by Aviva Plc. AIAR acknowledges the CMIH and CCIMI, University of Cambridge. CBS acknowledges the Philip Leverhulme, the EPSRC grants EP/S026045/1 and EP/T003553/1, EP/N014588/1, EP/T017961/1, NoMADS 777826, the CCIMI, University of Cambridge.

% HP acknowledges the financial support by Aviva Plc. AIAR gratefully acknowledges the financial support of the CMIH and CCIMI, University of Cambridge. CBS acknowledges support from the Philip Leverhulme Prize, the Royal Society Wolfson Fellowship, the EPSRC grants EP/S026045/1 and EP/T003553/1, EP/N014588/1, EP/T017961/1, the Wellcome Innovator Award RG98755, the Leverhulme Trust project Unveiling the invisible, the European Union Horizon 2020 research and innovation programme under the Marie Skodowska-Curie grant agreement No.\ 777826 NoMADS, the CCIMI and the Alan Turing Institute.

%------------------------------------------------------------------------
%{\small
%	\bibliographystyle{ieee_fullname}
%	\bibliography{egbib}
%}

\input{0094.bbl}
\newpage

\input{supp.tex}

\end{document}

%% file: supp.tex
\twocolumn[\centering \section*{\Large Supplementary Materials for HERS Superpixels: Deep Affinity Learning for Hierarchical Entropy Rate Segmentation}]
%\section{Outline} 
%\Angie{@Hankui: Since we have more sections, shall we keep Outline section for this part of the text?}
This document extends the network design details and visual results presented in the main paper, which is structured as follows.
\begin{itemize}[noitemsep]
    \item \textbf{Section~\ref{net_design}: Network Design.} We provide a detailed breakdown of the design of our proposed Deep Affinity Learning (DAL) network.
    \item \textbf{Section~\ref{boruvka_algo}: Further Details on Bor\r{u}vka's Algorithm.} We provide an explanation along with a visual example on the advantages of using Bor\r{u}vka's Algorithm.
	\item \textbf{Section~\ref{perf_measures}: Performance Measures.} In the interest of completeness, we detail the definitions of the performance measures that are used in the main paper to evaluate the superpixel segmentation results of various methods. 
	\item \textbf{Section~\ref{perf_results}: Supplementary Qualitative Results.} We showcase various additional visual comparisons: i) demonstrating the advantage of our method against other state-of-the-art superpixel methods; and ii) showcasing the adaptiveness of our segmentation results with varying numbers of user-specified superpixel counts.
	\item \textbf{Section~\ref{sal_detect}: Superpixels for saliency detection.} We apply our proposed DAL-HERS technique as a preprocessing step to the task of saliency detection. 
\end{itemize}

\section{Network Design}\label{net_design}
We provide further network design details of our Deep Affinity Learning (DAL) network in Table~\ref{tab_dal_details} and Table~\ref{tab_hed_details}.
We consider the setting where the given input image is of size $H\times W$, where $H=480$ and $W=320$. 
Each layer within a side output block is interleaved by a ReLU layer, which we omit in the table for ease of presentation. 

%% Explain the intermediate affinity map (produced from the first stage)
In the first part of our DAL network, we obtain an intermediate affinity map (\textit{out3c}) at the end of the three ResBlocks. This is used as input to the second stage, which mainly consists of the HED network structure (see Table~\ref{tab_hed_details}) for further boundary information learning.
%% Explain the HED network 
% How are the side outputs fused together 
Within the HED structure, the five side outputs as obtained at the end of each of the five Bilinear Interpolation (BI) steps all have size $8\times 480\times 320$. They are concatenated together to form a tensor of size $40\times 480\times 320$, which serves as input to a convolutional layer (fusion layer) that converts 40 channels down to 8 channels. Finally, the learned affinity map is obtained by applying the Sigmoid function to the output of the aforementioned fusion layer. 

\section{Further Details on Bor\r{u}vka's Algorithm}\label{boruvka_algo}
\begin{figure}[htbp!]
	\centering
	\begin{subfigure}[b]{.32\linewidth}
		\includegraphics[width=\linewidth, height=.8\linewidth]{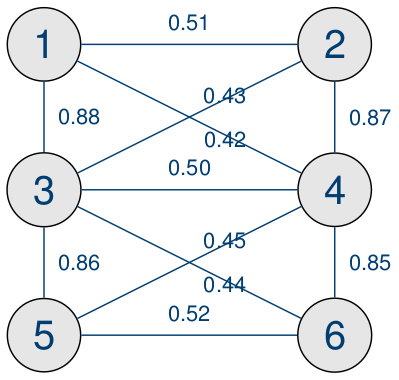}
		\caption{}
	\end{subfigure}
	\begin{subfigure}[b]{.32\linewidth}
		\includegraphics[width=\linewidth, height=.8\linewidth]{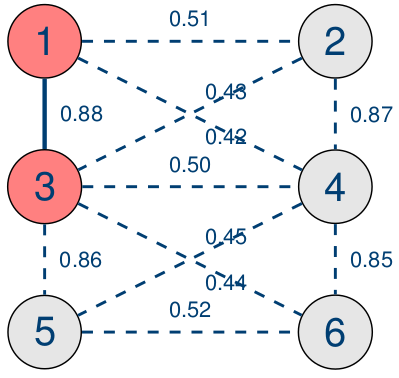}
		\caption{}
	\end{subfigure}
	\begin{subfigure}[b]{.32\linewidth}
		\includegraphics[width=\linewidth, height=.8\linewidth]{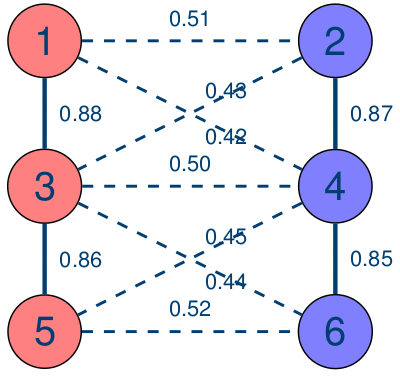}
		\caption{}
	\end{subfigure}
	\caption{A simple example that illustrates the efficiency and parallelisability of Bor\r{u}vka's algorithm. (a) displays an initial graph containing six nodes and their weighted edges.  Bor\r{u}vka's algorithm,  displayed in (c), only needs one iteration to arrive at the final partitioning, whereas the lazy greedy algorithm~\cite{liu2011entropy} in (b) requires four iterations.} 
	\label{fig::Borruvka}
\end{figure}

In the main paper, we present a graph-based framework which consists of a neural network for deep affinity learning, and an efficient superpixel segmentation method HERS for obtaining highly adaptive superpixels efficiently. An illustration of HERS is displayed in Figure~\ref{fig::Borruvka}. It is clear that Bor\r{u}vka's algorithm already arrived at the correct segmentation at the end of iteration 1, whereas the lazy greedy algorithm has only connected two nodes together.   

\begin{table*}[htbp!]
\centering
\begin{tabular}{|c|c|c|c|c|c|c|c|}
\hline
\multirow{2}{*}{{Operation}} &\multirow{2}{1cm}{\centering Input} & \multirow{2}{1cm}{\centering Output} & \multirow{2}{1cm}{\centering Kernel size} & \multirow{2}{1cm}{\centering Stride size} & \multirow{2}{1cm}{\centering Channel I/O} & \multirow{2}{*}{\centering Input Res.} & \multirow{2}{*}{\centering Output Res.} \\ 
&&&&&&&\\
\hline
Conv. &\multicolumn{1}{|c|}{image} & \multicolumn{1}{c|}{out1} & \multicolumn{1}{c|}{7} & \multicolumn{1}{c|}{1} & \multicolumn{1}{c|}{3/8} &$3\times 480\times 320$&$8\times 480\times 320$\\ 
\hline
Ins.\ Norm. &\multicolumn{1}{|c|}{out1} & \multicolumn{1}{c|}{out2} & \multicolumn{1}{c|}{-}     & \multicolumn{1}{c|}{-} & \multicolumn{1}{c|}{8/8} &$8\times 480\times 320$&$8\times 480\times 320$ \\ 
\hline
Relu &out2&out3&-&-&8/8&$8\times 480\times 320$&$8\times 480\times 320$\\
\hline 
ResBlock & out3 & out3a & 3&1 & 8/8&$8\times 480\times 320$&$8\times 480\times 320$\\ 
\hline
ResBlock & out3a & out3b & 3&1 & 8/8&$8\times 480\times 320$&$8\times 480\times 320$\\ 
\hline
ResBlock & out3b & out3c & 3&1 & 8/8&$8\times 480\times 320$&$8\times 480\times 320$\\ 
\hline
HED &out3c & out &\multicolumn{5}{|c|}{}\\
\hline 
\end{tabular}
\caption{Specification of the proposed Deep Affinity Learning (DAL) network structure.}
\label{tab_dal_details}
\end{table*}

\begin{table*}[htbp!]
	\centering
	\begin{tabular}{|c|c|c|c|c|c|c|c|}
		\hline
		\multirow{2}{*}{Operation} &\multirow{2}{1cm}{\centering Input} & \multirow{2}{1cm}{\centering Output} & \multirow{2}{1cm}{\centering Kernel size} & \multirow{2}{1cm}{\centering Stride size} & \multirow{2}{1cm}{\centering Channel I/O} & \multirow{2}{*}{\centering Input Res.} & \multirow{2}{*}{\centering Output Res.} \\ 
		&&&&&&&\\
		%%%%%%%%%%%%%%%%%%%%%%%%%%%%%%%
		\hline
		\multirow{2}{*}{Side Output 1} & \multicolumn{1}{|c|}{out3c} & \multicolumn{1}{c|}{hed1a} & \multicolumn{1}{c|}{3} & \multicolumn{1}{c|}{1} & \multicolumn{1}{c|}{8/64} &$8\times 480\times 320$&$64\times 480\times 320$\\ 
		& \multicolumn{1}{|c|}{hed1a} & \multicolumn{1}{c|}{hed1} & \multicolumn{1}{c|}{3} & \multicolumn{1}{c|}{1} & \multicolumn{1}{c|}{64/64} &$64\times 486\times 326$ & $64\times 480\times 320$\\ 
		\hline
		%%%%%%%%%%%%%%%%%%%%%%%%%%%%%%%
		\multirow{3}{*}{Side Output 2} & \multicolumn{7}{|c|}{Max Pooling (Kernel size $= 2$, Stride $= 2$) } \\
		& \multicolumn{1}{|c|}{hed1} & \multicolumn{1}{c|}{hed2a} & \multicolumn{1}{c|}{3} & \multicolumn{1}{c|}{1} & \multicolumn{1}{c|}{64/128} &$64\times 240\times 160$&$128\times 240\times 160$\\ 
		& \multicolumn{1}{|c|}{hed2a} & \multicolumn{1}{c|}{hed2} & \multicolumn{1}{c|}{3} & \multicolumn{1}{c|}{1} & \multicolumn{1}{c|}{128/128} &$128\times 240\times 160$ & $128\times 240\times 160$\\ 
		\hline 
		%%%%%%%%%%%%%%%%%%%%%%%%%%%%%%%
		\multirow{4}{*}{Side Output 3} & \multicolumn{7}{|c|}{Max Pooling (Kernel size $= 2$, Stride $= 2$) } \\
		& \multicolumn{1}{|c|}{hed2} & \multicolumn{1}{c|}{hed3a} & \multicolumn{1}{c|}{3} & \multicolumn{1}{c|}{1} & \multicolumn{1}{c|}{128/256} &$128\times 120\times 80$&$256\times 120\times 80$\\ 
		& \multicolumn{1}{|c|}{hed3a} & \multicolumn{1}{c|}{hed3b} & \multicolumn{1}{c|}{3} & \multicolumn{1}{c|}{1} & \multicolumn{1}{c|}{256/256} &$256\times 120\times 80$ & $256\times 120\times 80$\\ 
		& \multicolumn{1}{|c|}{hed3b} & \multicolumn{1}{c|}{hed3}& \multicolumn{1}{c|}{3} & \multicolumn{1}{c|}{1} & \multicolumn{1}{c|}{256/256} &$256\times 120\times 80$ & $256\times 120\times 80$\\ 
		\hline 
		%%%%%%%%%%%%%%%%%%%%%%%%%%%%%%%
		\multirow{4}{*}{Side Output 4} & \multicolumn{7}{|c|}{Max Pooling (Kernel size $= 2$, Stride $= 2$) } \\
		& \multicolumn{1}{|c|}{hed3} & \multicolumn{1}{c|}{hed4a} & \multicolumn{1}{c|}{3} & \multicolumn{1}{c|}{1} & \multicolumn{1}{c|}{256/512} &$256\times 60\times 40$&$512\times 60\times 40$\\ 
		& \multicolumn{1}{|c|}{hed4a} & \multicolumn{1}{c|}{hed4b} & \multicolumn{1}{c|}{3} & \multicolumn{1}{c|}{1} & \multicolumn{1}{c|}{512/512} &$512\times 60\times 40$ & $512\times 60\times 40$\\ 
		& \multicolumn{1}{|c|}{hed4b} & \multicolumn{1}{c|}{hed4}& \multicolumn{1}{c|}{3} & \multicolumn{1}{c|}{1} & \multicolumn{1}{c|}{512/512} &$512\times 60\times 40$ & $512\times 60\times 40$\\ 
		\hline  
		%%%%%%%%%%%%%%%%%%%%%%%%%%%%%%%
		\multirow{4}{*}{Side Output 5} & \multicolumn{7}{|c|}{Max Pooling (Kernel size $= 2$, Stride $= 2$) } \\
		& \multicolumn{1}{|c|}{hed4} & \multicolumn{1}{c|}{hed5a} & \multicolumn{1}{c|}{3} & \multicolumn{1}{c|}{1} & \multicolumn{1}{c|}{256/512} &$512\times 30\times 20$&$512\times 30\times 20$\\ 
		& \multicolumn{1}{|c|}{hed5a} & \multicolumn{1}{c|}{hed5b} & \multicolumn{1}{c|}{3} & \multicolumn{1}{c|}{1} & \multicolumn{1}{c|}{512/512} &$512\times 30\times 20$ & $512\times 30\times 20$\\ 
		& \multicolumn{1}{|c|}{hed5b} & \multicolumn{1}{c|}{hed5}& \multicolumn{1}{c|}{3} & \multicolumn{1}{c|}{1} & \multicolumn{1}{c|}{512/512} &$512\times 30\times 20$ & $512\times 30\times 20$\\ 
		\hline 
		\hline 
		\multirow{2}{*}{Conv. 1} &hed1 &hed1\_out & 1&1&64/8&$64\times 480\times 320$ & $8\times 480\times 320$\\
		&\multicolumn{7}{c|}{Bilinear Interpolation (BI), Output size $=8\times 480\times 320$}\\
		\hline 
		\multirow{2}{*}{Conv. 2} &hed2 &hed2\_out & 1&1&128/8&$128\times 240\times 160$ & $8\times 240\times 160$\\
		&\multicolumn{7}{c|}{Bilinear Interpolation (BI), Output size $=8\times 480\times 320$}\\
		\hline 
		\multirow{2}{*}{Conv. 3} &hed3 &hed3\_out & 1&1&256/8&$256\times 120\times 80$ & $8\times 120\times 80$\\
		&\multicolumn{7}{c|}{Bilinear Interpolation (BI), Output size $=8\times 480\times 320$}\\
		\hline
		\multirow{2}{*}{Conv. 4} &hed4 &hed4\_out & 1&1&512/8&$512\times 60\times 40$ & $8\times 60\times 40$\\
		&\multicolumn{7}{c|}{Bilinear Interpolation (BI), Output size $=8\times 480\times 320$}\\
		\hline
		\multirow{2}{*}{Conv. 5} &hed5 &hed5\_out & 1&1&512/8&$512\times 30\times 20$ & $8\times 30\times 20$\\
		&\multicolumn{7}{c|}{Bilinear Interpolation (BI), Output size $=8\times 480\times 320$}\\
		\hline
		\multirow{2}{2.3cm}{\centering Conv. \\(Fusion Layer)} & \multirow{2}{*}{5 BI outputs}&\multirow{2}{*}{combined\_out}&\multirow{2}{*}{1}&\multirow{2}{*}{1}&\multirow{2}{*}{40/8}&\multirow{2}{*}{$8\times 480\times 320$}&\multirow{2}{*}{$8\times 480\times 320$}\\
		&&&&&&&\\
		\hline 
	\end{tabular}
	\caption{Specification of the HED component within the DAL network.}
\label{tab_hed_details}
\end{table*}

\section{Performance Measures}~\label{perf_measures}
The performance of various superpixel segmentation algorithms are commonly measured by the Under-segmentation Error (UE)~\cite{van2012seeds}, Achievable Segmentation Accuracy (ASA)~\cite{liu2011entropy} and Boundary Recall (BR)~\cite{martin2004learning}. UE compares each computed superpixel with the ground truth superpixel that it overlaps with the most, and measures the ``leakage'' area that are not in the overlapped region. Opposite to UE, ASA quantifies the percentage of overlap between the segmented superpixels and the ground truth superpixels. That is, ASA can be directly obtained from UE as ASA=1-UE. 

As such, we choose one out of these two and report ASA in our experiments. Concretely, ASA can be computed as  
\begin{equation}
\text{ASA}(\gtSeg, \spSeg)=\frac{1}{N}\sum_{k=1}^{\ncluster}\underset{\gtSeg_{c}}{\arg\max} |\spSeg_{k}\cap\gtSeg_{c}|,
\end{equation}
in which $\gtSeg=\left\{\gtSeg_{1},\ldots,\gtSeg_{\nclass} \right\}$ denotes the ground truth segmentation, and $\spSeg=\left\{\spSeg_{1},\ldots,\spSeg_{\ncluster} \right\}$ denotes the segmentation given by the chosen algorithm.

Boundary Recall (BR) measures the boundary adherence of the computed superpixels to the ground truth boundaries. It measures the proportion of ground truth boundary pixels that have been correctly identified by the computed superpixels. Concretely, BR can be computed as 
\begin{equation}
\text{BR}(\gtSeg, \spSeg)=\frac{\text{TP}(\gtSeg, \spSeg)}{\text{TP}(\gtSeg, \spSeg)+\text{FN}(\gtSeg, \spSeg)},
\end{equation}
in which $\text{TP}(\gtSeg, \spSeg)$ stands for true positive, it denotes the number of ground truth boundary pixels that have been identified by the superpixel segmentation algorithm. $\text{FN}(\gtSeg, \spSeg)$ stands for false negative, which denotes the remaining number of ground truth boundary pixels that have not been identified.

Additionally, we also report the Explained Variation (EV)~\cite{moore2008superpixel} score, which quantifies the variance within an image that is captured by the superpixels without relying on any ground truth labelling. It is calculated using the following formula
\begin{equation}
\text{EV}(\spSeg)=\frac{\sum_{i} \left(\boldsymbol{\mu}_{i} -\boldsymbol{\mu} \right)^{2}}{\sum_{i}\left(\pixel_{i} -\boldsymbol{\mu} \right)^{2}},
\end{equation}
where $\pixel_{i}$ denotes the RGB pixel values for the $i$-th pixel, $\boldsymbol{\mu}$ denotes the global mean of the RGB pixel values of an image, and $\boldsymbol{\mu}_{i}$ denotes the mean RGB pixel values for the superpixel that contains pixel $\pixel_{i}$.

\section{Supplementary Qualitative Results}\label{perf_results}
In this section, we provide further visual comparisons on the BSDS500~\cite{arbelaez2010contour} and NYUv2~\cite{silberman2012indoor} datasets across the following state-of-the-art techniques: i) classic techniques: ERS~\cite{liu2011entropy}, SH~\cite{wei2018superpixel}, SLIC~\cite{achanta2012slic}, SNIC~\cite{achanta2017superpixels},  SEEDS~\cite{van2012seeds}, ETPS~\cite{yao2015real}; and ii) deep learning techniques:  SSN~\cite{jampani2018superpixel}, SEAL-ERS~\cite{tu2018learning} and SP-FCN~\cite{yang2020superpixel}. 

\subsection{Additional visual comparisons against state-of-the-art methods}
\begin{figure*}[ht!]
\centering
\begin{subfigure}[b]{.33\textwidth}
	%% original 
	\includegraphics[width=\linewidth, height=.7\linewidth]{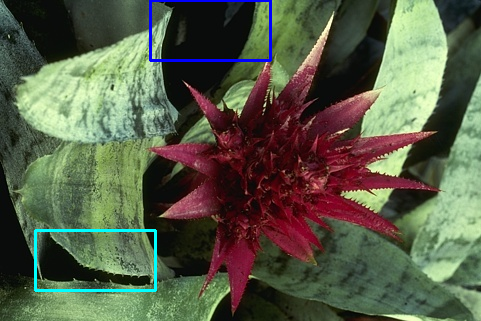}
	\includegraphics[width=.485\linewidth]{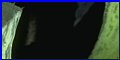}
	\includegraphics[width=.485\linewidth]{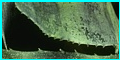}
	\caption{Original.}
\end{subfigure}
\begin{subfigure}[b]{.33\textwidth}
	\includegraphics[width=\linewidth, height=.7\linewidth]{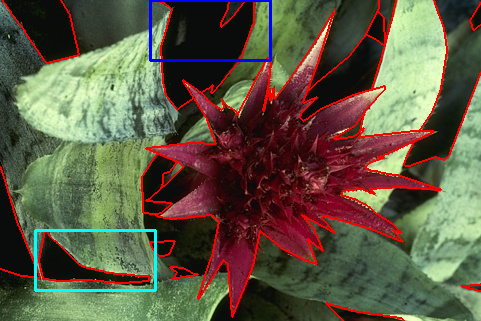}
	\includegraphics[width=.485\linewidth]{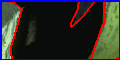}
	\includegraphics[width=.485\linewidth]{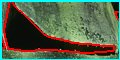}
	\caption{Ground truth.}
\end{subfigure}
\begin{subfigure}[b]{.33\textwidth}
	%% ours 
	\includegraphics[width=\linewidth, height=.7\linewidth]{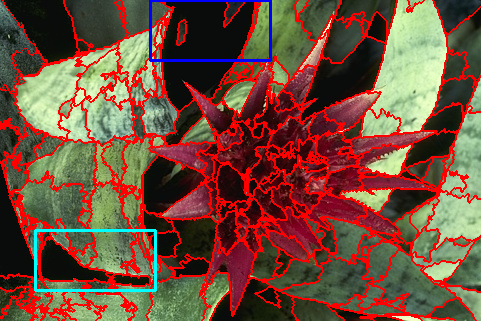}
	\includegraphics[width=.485\linewidth]{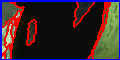}
	\includegraphics[width=.485\linewidth]{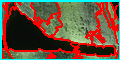}
	\caption{Ours.}
\end{subfigure}
\begin{subfigure}[b]{.33\textwidth}
	%% ers
	\includegraphics[width=\linewidth, height=.7\linewidth]{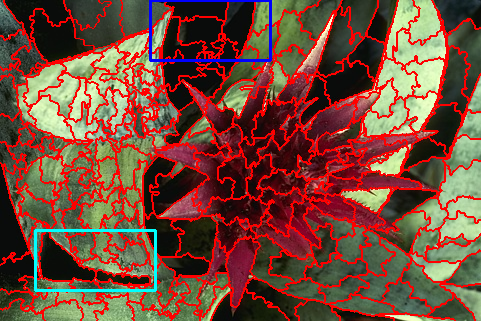}
	\includegraphics[width=.485\linewidth]{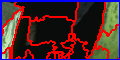}
	\includegraphics[width=.485\linewidth]{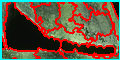}
	\caption{ERS.}
\end{subfigure}
\begin{subfigure}[b]{.33\textwidth}
	%% sh
	\includegraphics[width=\linewidth, height=.7\linewidth]{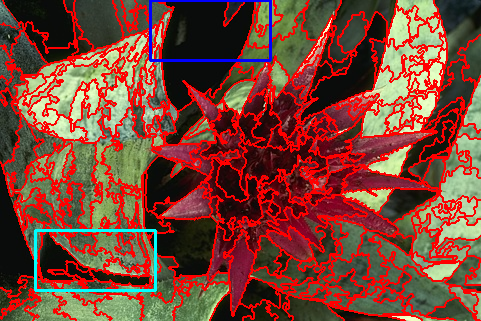}
	\includegraphics[width=.485\linewidth]{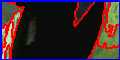}
	\includegraphics[width=.485\linewidth]{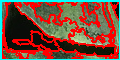}
	\caption{SH.}
\end{subfigure}
\begin{subfigure}[b]{.33\textwidth}
	%% slic
	\includegraphics[width=\linewidth, height=.7\linewidth]{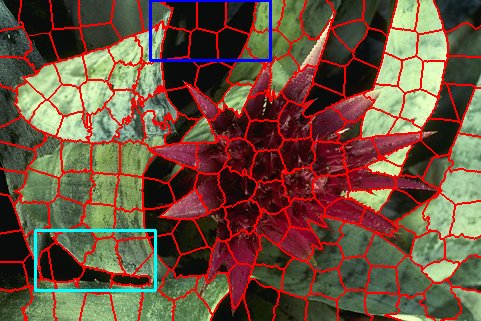}
	\includegraphics[width=.485\linewidth]{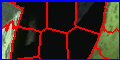}
	\includegraphics[width=.485\linewidth]{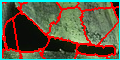}
	\caption{SLIC.}
\end{subfigure}
\begin{subfigure}[b]{.33\textwidth}
	%% snic
	\includegraphics[width=\linewidth, height=.7\linewidth]{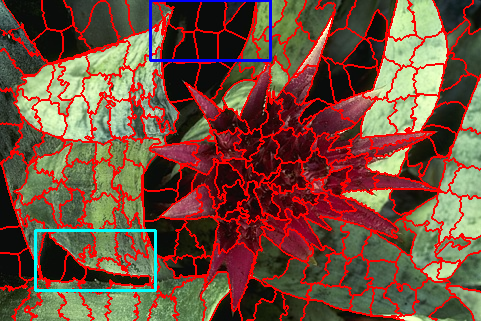}
	\includegraphics[width=.485\linewidth]{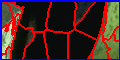}
	\includegraphics[width=.485\linewidth]{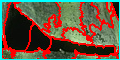}
	\caption{SNIC.}
\end{subfigure}
\begin{subfigure}[b]{.33\textwidth}
	%% seeds
	\includegraphics[width=\linewidth, height=.7\linewidth]{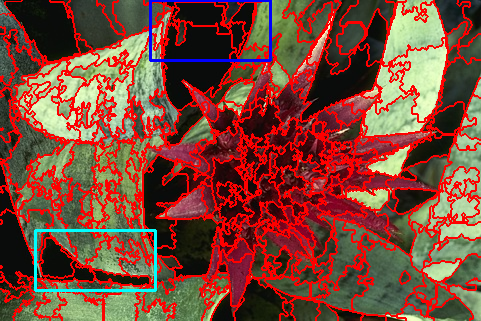}
	\includegraphics[width=.485\linewidth]{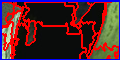}
	\includegraphics[width=.485\linewidth]{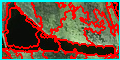}
	\caption{SEEDS.}
\end{subfigure}
\begin{subfigure}[b]{.33\textwidth}
	%% etps
	\includegraphics[width=\linewidth, height=.7\linewidth]{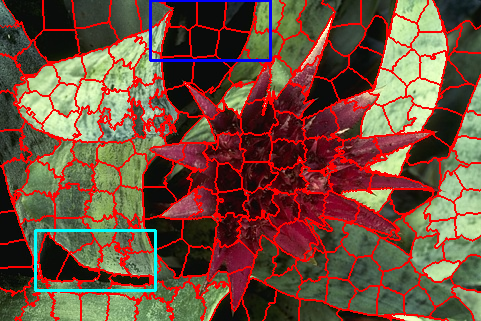}
	\includegraphics[width=.485\linewidth]{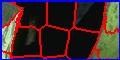}
	\includegraphics[width=.485\linewidth]{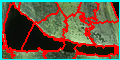}
	\caption{ETPS.}
\end{subfigure}
\begin{subfigure}[b]{.33\textwidth}
	%% seal
	\includegraphics[width=\linewidth, height=.7\linewidth]{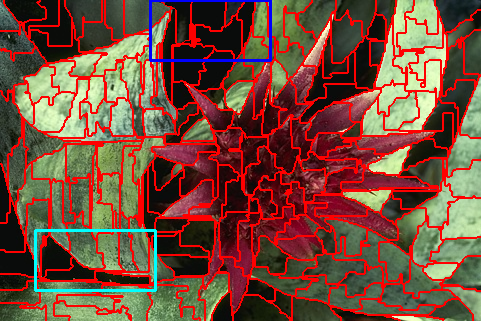}
	\includegraphics[width=.485\linewidth]{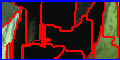}
	\includegraphics[width=.485\linewidth]{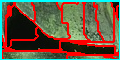}
	\caption{SEAL-ERS.}
\end{subfigure}
\begin{subfigure}[b]{.33\textwidth}
	%% SP-FCN 
	\includegraphics[width=\linewidth, height=.7\linewidth]{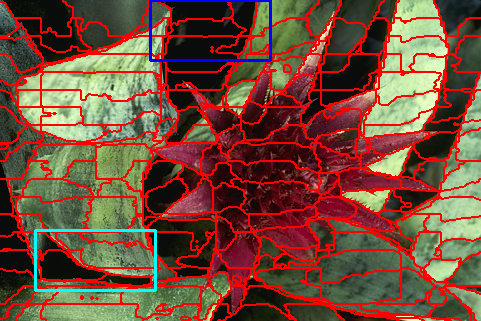}
	\includegraphics[width=.485\linewidth]{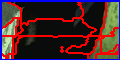}
	\includegraphics[width=.485\linewidth]{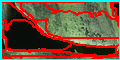}
	\caption{SP-FCN.}										
\end{subfigure}
\begin{subfigure}[b]{.33\textwidth}
	%% SSN 
	\includegraphics[width=\linewidth, height=.7\linewidth]{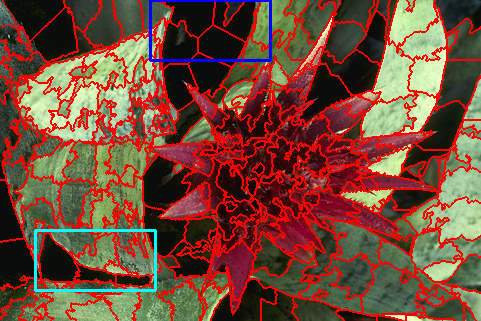}
	\includegraphics[width=.485\linewidth]{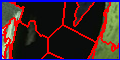}
	\includegraphics[width=.485\linewidth]{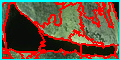}
	\caption{SSN.}
\end{subfigure}
\caption{Segmentation results on a sample image from the BSDS500 test set with 200 superpixels.}
\label{fig:bsds1}
\end{figure*}

%Figure~\ref{fig:bsds1} and \ref{fig:bsds2} present additional visual comparisons of various methods on the BSDS500 dataset. Figure~\ref{fig:nyu1} and \ref{fig:nyu2} contain additional visual comparisons of various methods on the NYUv2 dataset.
Figures~\ref{fig:bsds1} %and \ref{fig:bsds2} display two supplementary visual examples 
displays an additional example 
from the BSDS500 dataset. We note that amongst the compared techniques, our technique presents the most visually appealing output. By visual inspection, we can notice that  our superpixels clearly show better segmentations of fine details and stronger boundary adherence whilst avoiding partitioning homogeneous regions. In particular, several methods including the deep learning techniques of SEAL-ERS (see output (j)) and SP-FCN (see output (k)) often fail to capture the boundary structures. 
By contrast, our superpixels are better at capturing the objects of the scene including complex ones such as the shape of the flower in Figures~\ref{fig:bsds1}. % and \ref{fig:bsds2}. 
Some examples of these advantages are highlighted in the zoomed-in views. 

\begin{figure*}[ht!]
\centering
\begin{subfigure}[b]{.33\textwidth}
	%% original 
	\includegraphics[width=\linewidth, height=.7\linewidth]{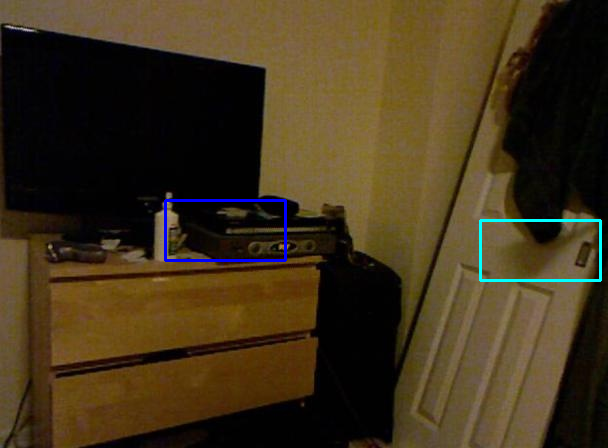}
	\includegraphics[width=.485\linewidth]{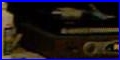}
	\includegraphics[width=.485\linewidth]{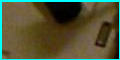}
	\caption{Original.}
\end{subfigure}
\begin{subfigure}[b]{.33\textwidth}
	\includegraphics[width=\linewidth, height=.7\linewidth]{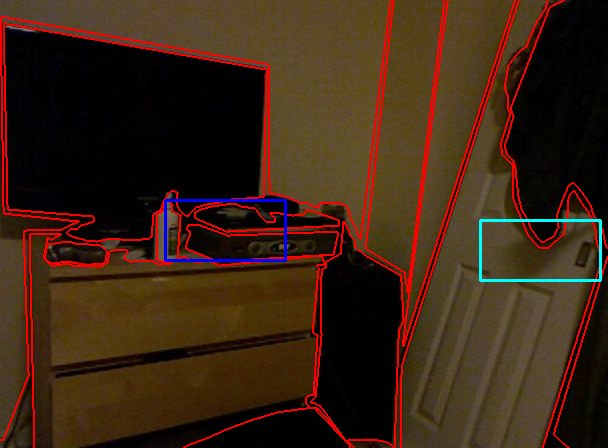}
	\includegraphics[width=.485\linewidth]{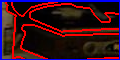}
	\includegraphics[width=.485\linewidth]{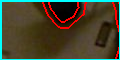}
	\caption{Ground truth.}
\end{subfigure}
\begin{subfigure}[b]{.33\textwidth}
	%% ours 
	\includegraphics[width=\linewidth, height=.7\linewidth]{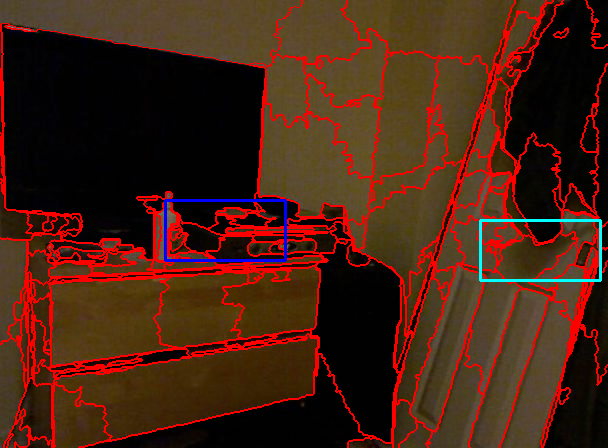}
	\includegraphics[width=.485\linewidth]{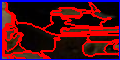}
	\includegraphics[width=.485\linewidth]{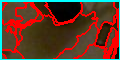}
	\caption{Ours.}
\end{subfigure}
\begin{subfigure}[b]{.33\textwidth}
	%% ers
	\includegraphics[width=\linewidth, height=.7\linewidth]{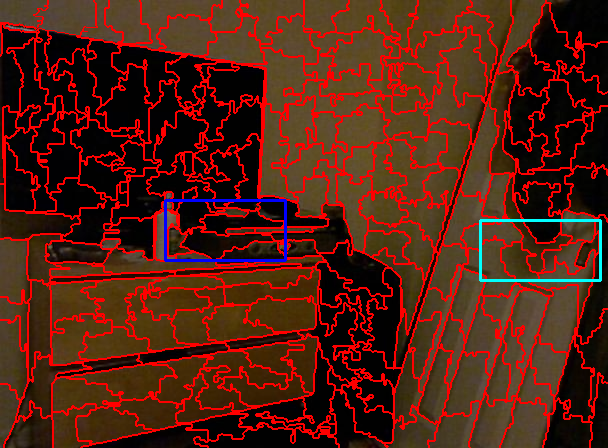}
	\includegraphics[width=.485\linewidth]{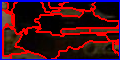}
	\includegraphics[width=.485\linewidth]{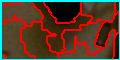}
	\caption{ERS.}
\end{subfigure}
\begin{subfigure}[b]{.33\textwidth}
	%% sh
	\includegraphics[width=\linewidth, height=.7\linewidth]{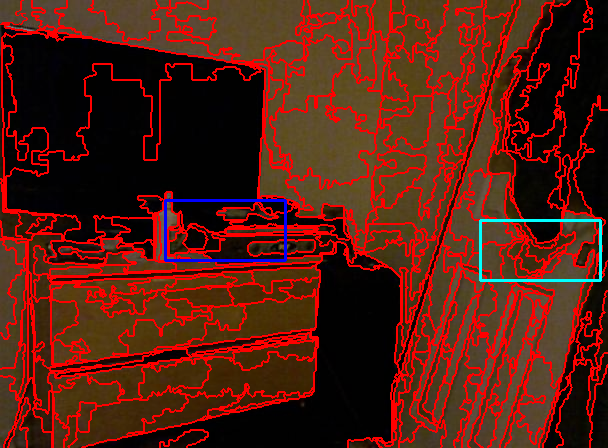}
	\includegraphics[width=.485\linewidth]{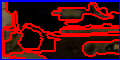}
	\includegraphics[width=.485\linewidth]{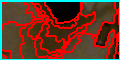}
	\caption{SH.}
\end{subfigure}
\begin{subfigure}[b]{.33\textwidth}
	%% slic
	\includegraphics[width=\linewidth, height=.7\linewidth]{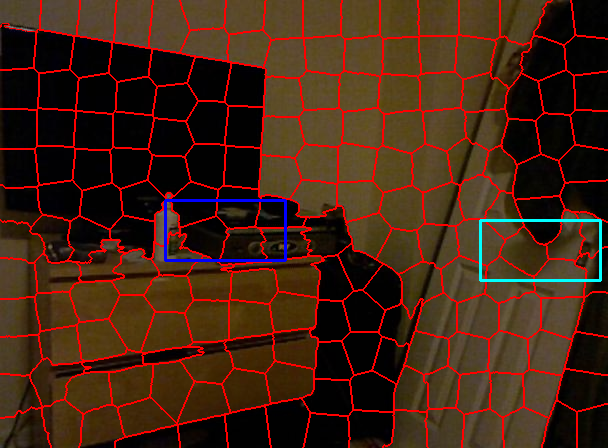}
	\includegraphics[width=.485\linewidth]{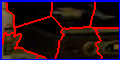}
	\includegraphics[width=.485\linewidth]{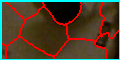}
	\caption{SLIC.}
\end{subfigure}
\begin{subfigure}[b]{.33\textwidth}
	%% snic
	\includegraphics[width=\linewidth, height=.7\linewidth]{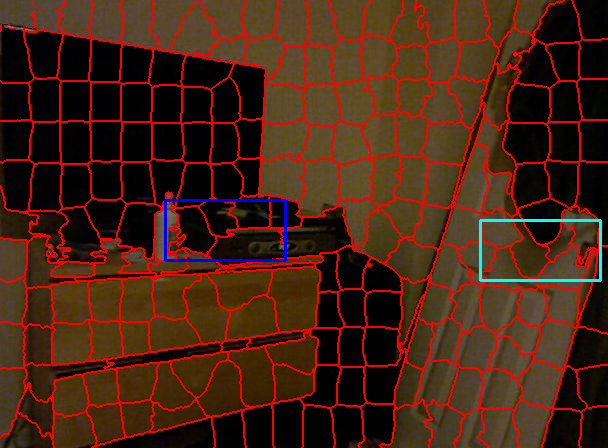}
	\includegraphics[width=.485\linewidth]{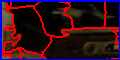}
	\includegraphics[width=.485\linewidth]{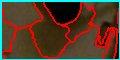}
	\caption{SNIC.}
\end{subfigure}
\begin{subfigure}[b]{.33\textwidth}
	%% seeds
	\includegraphics[width=\linewidth, height=.7\linewidth]{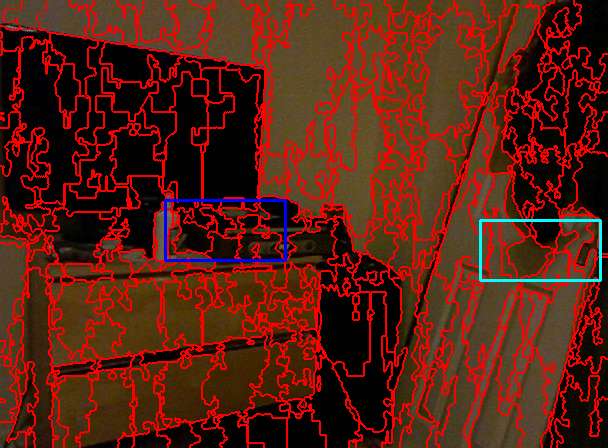}
	\includegraphics[width=.485\linewidth]{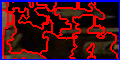}
	\includegraphics[width=.485\linewidth]{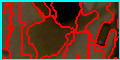}
	\caption{SEEDS.}
\end{subfigure}
\begin{subfigure}[b]{.33\textwidth}
	%% etps
	\includegraphics[width=\linewidth, height=.7\linewidth]{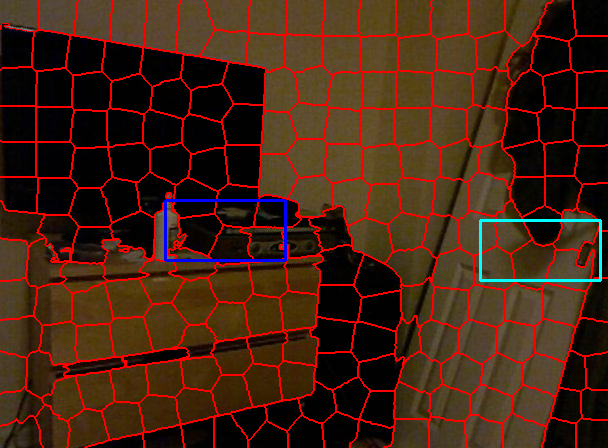}
	\includegraphics[width=.485\linewidth]{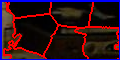}
	\includegraphics[width=.485\linewidth]{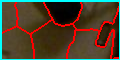}
	\caption{ETPS.}
\end{subfigure}
\begin{subfigure}[b]{.33\textwidth}
	%% seal
	\includegraphics[width=\linewidth, height=.7\linewidth]{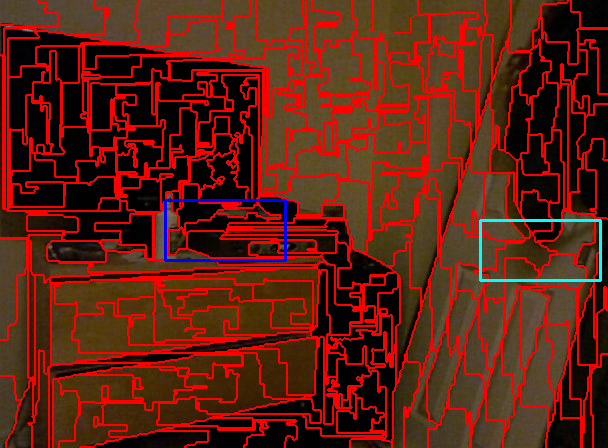}
	\includegraphics[width=.485\linewidth]{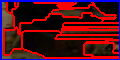}
	\includegraphics[width=.485\linewidth]{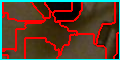}
	\caption{SEAL-ERS.}
\end{subfigure}
\begin{subfigure}[b]{.33\textwidth}
	%% SP-FCN 
	\includegraphics[width=\linewidth, height=.7\linewidth]{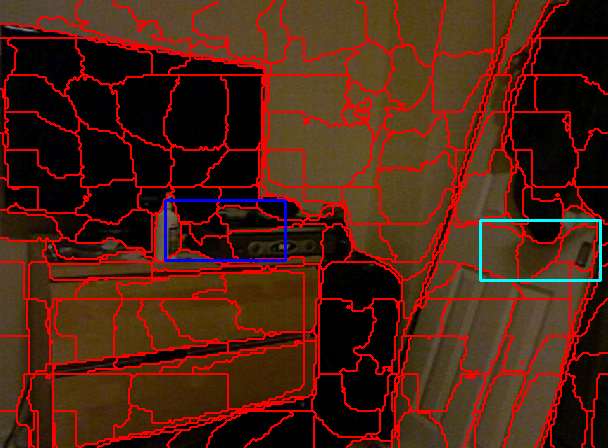}
	\includegraphics[width=.485\linewidth]{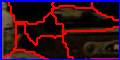}
	\includegraphics[width=.485\linewidth]{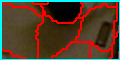}
	\caption{SP-FCN.}										
\end{subfigure}
\begin{subfigure}[b]{.33\textwidth}
	%% SSN 
	\includegraphics[width=\linewidth, height=.7\linewidth]{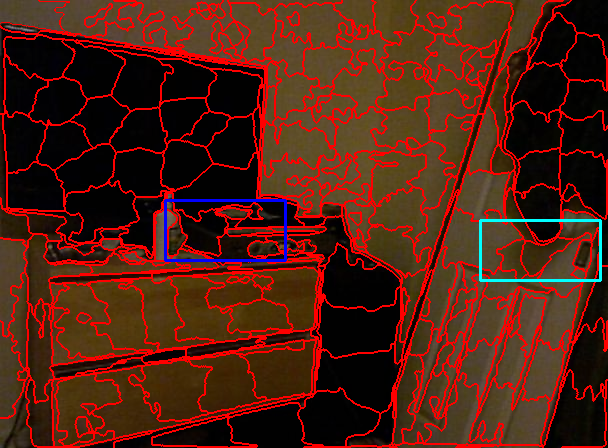}
	\includegraphics[width=.485\linewidth]{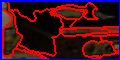}
	\includegraphics[width=.485\linewidth]{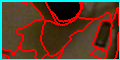}
	\caption{SSN.}
\end{subfigure}
\caption{Segmentation results on a sample image from the NYUv2 test set with 200 superpixels.}
\label{fig:nyu1}
\end{figure*}

These benefits of our technique are also observed in indoors scenes as displayed in Figures~\ref{fig:nyu1}, % and \ref{fig:nyu2}, 
which are taken from the NYUv2 test set. We selected interesting samples with complex objects of varying sizes. 
In Figure~\ref{fig:nyu1}, one can observe that none of the competing techniques are able to adhere well to the boundaries of small objects (e.g.\ see the highlighted part in the blue zoomed-in square). 
% This advantage of ours is expected since our technique avoids partitioning homogeneous areas (e,g,\ see the TV) and focuses on capturing the fine details in texture-rich scenes. This effect is even more noticeable in Figure~\ref{fig:nyu2}, where our technique delineates the details of the main objects. Some highlighted cases can be seen in the zoomed-in views.

\subsection{Various numbers of superpixels $K$}
\begin{figure*}
% line 1
\begin{subfigure}[b]{.33\textwidth}
\includegraphics[width=\linewidth, height=1.2\linewidth]{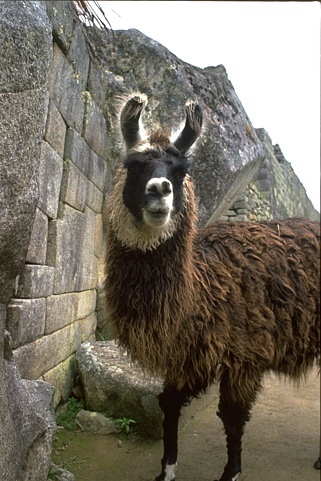}
\caption{Original image.}
\end{subfigure}
\begin{subfigure}[b]{.33\textwidth}
\includegraphics[width=\linewidth, height=1.2\linewidth]{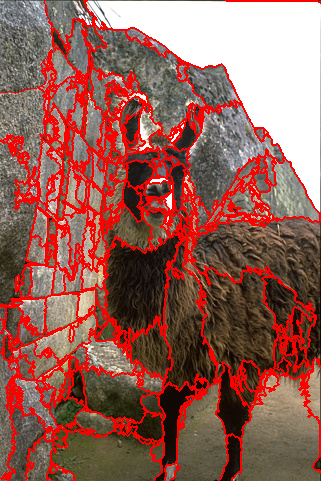}
\caption{$K=200$.}
\end{subfigure}
\begin{subfigure}[b]{.33\textwidth}
\includegraphics[width=\linewidth, height=1.2\linewidth]{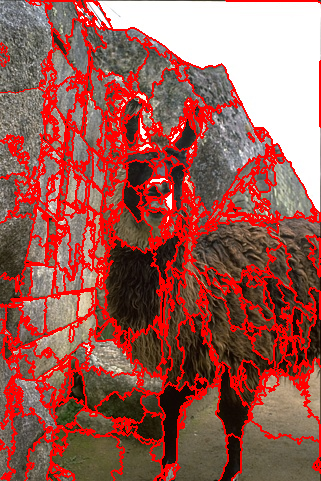}
\caption{$K=400$.}
\end{subfigure}
% line 2
\begin{subfigure}[b]{.33\textwidth}
	\includegraphics[width=\linewidth, height=1.2\linewidth]{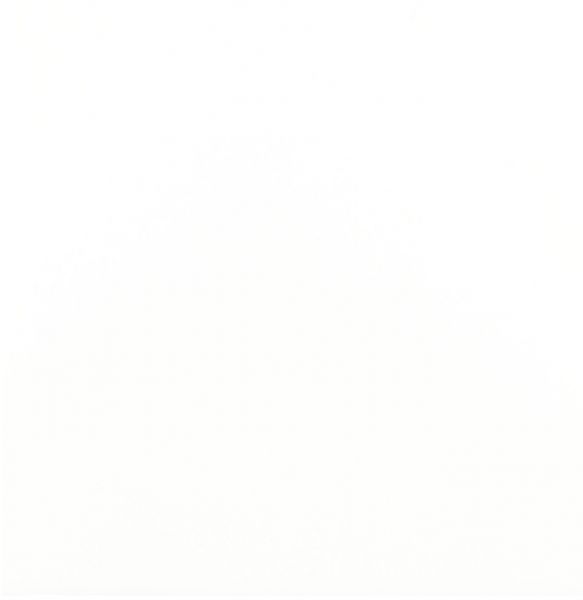}
\end{subfigure}
\begin{subfigure}[b]{.33\textwidth}
\includegraphics[width=\linewidth, height=1.2\linewidth]{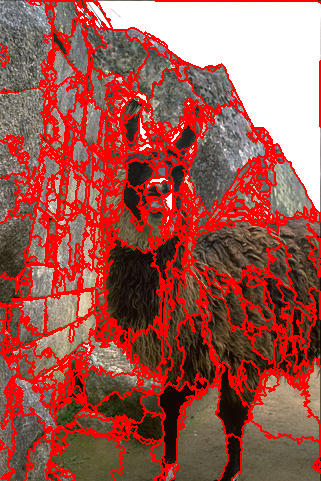}
\caption{$K=600$.}
\end{subfigure}
\begin{subfigure}[b]{.33\textwidth}
\includegraphics[width=\linewidth, height=1.2\linewidth]{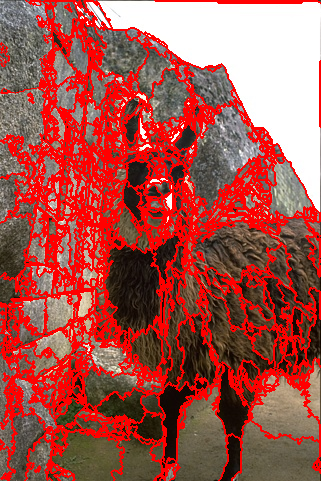}
\caption{$K=800$.}
\end{subfigure}
% line 3
\begin{subfigure}[b]{.33\textwidth}
\includegraphics[width=\linewidth, height=1.2\linewidth]{imgs/placeholder.jpeg}
\end{subfigure}
\begin{subfigure}[b]{.33\textwidth}
\includegraphics[width=\linewidth, height=1.2\linewidth]{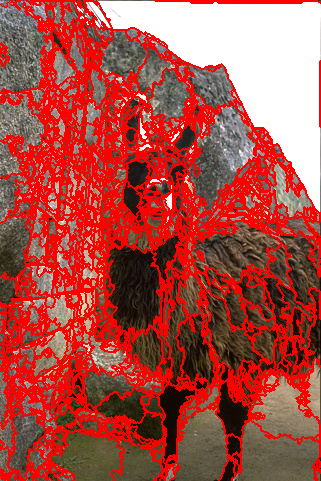}
\caption{$K=1000$.}
\end{subfigure}
\begin{subfigure}[b]{.33\textwidth}
\includegraphics[width=\linewidth, height=1.2\linewidth]{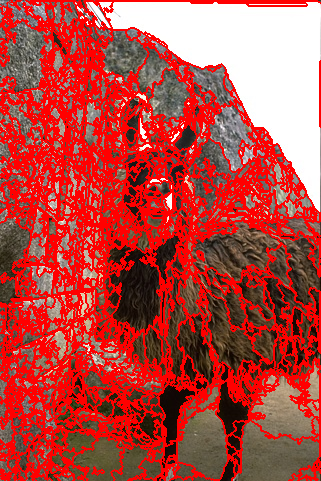}
\caption{$K=1200$.}
\end{subfigure}
\caption{Segmentation results using our method on an image from the BSDS500 test set with varying number of superpixels.}
\label{fig:varying_k_1a}
\end{figure*}

\begin{figure*}
% line 1
\begin{subfigure}[b]{.33\textwidth}
	\includegraphics[width=\linewidth, height=1.2\linewidth]{imgs/6046_img/6046_img.jpg}
	\caption{Original image.}
\end{subfigure}
\begin{subfigure}[b]{.33\textwidth}
	\includegraphics[width=\linewidth, height=1.2\linewidth]{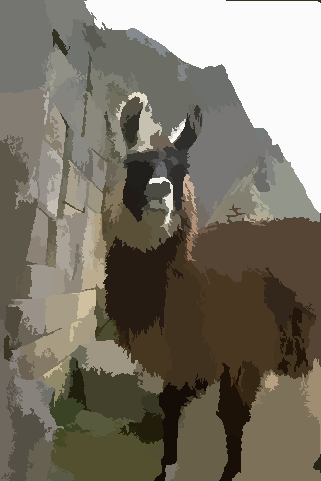}
	\caption{$K=200$.}
\end{subfigure}
\begin{subfigure}[b]{.33\textwidth}
	\includegraphics[width=\linewidth, height=1.2\linewidth]{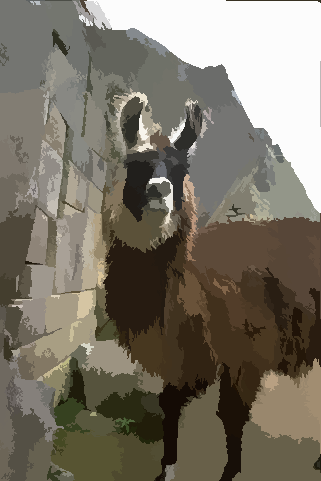}
	\caption{$K=400$.}
\end{subfigure}
% line 2
\begin{subfigure}[b]{.33\textwidth}
	\includegraphics[width=\linewidth, height=1.2\linewidth]{imgs/placeholder.jpeg}
\end{subfigure}
\begin{subfigure}[b]{.33\textwidth}
	\includegraphics[width=\linewidth, height=1.2\linewidth]{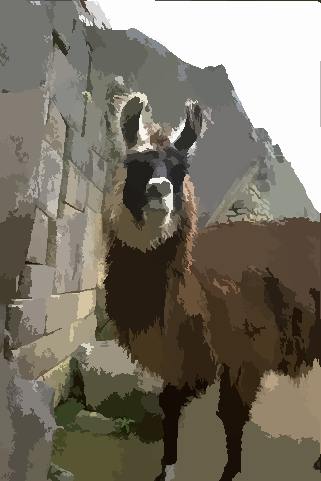}
	\caption{$K=600$.}
\end{subfigure}
\begin{subfigure}[b]{.33\textwidth}
	\includegraphics[width=\linewidth, height=1.2\linewidth]{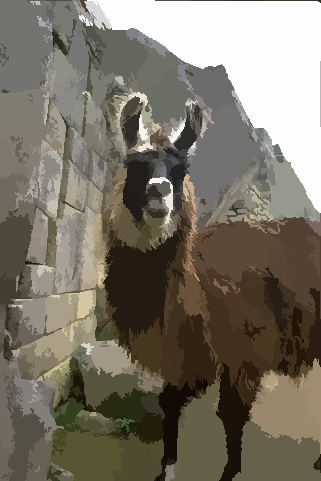}
	\caption{$K=800$.}
\end{subfigure}
% line 3
\begin{subfigure}[b]{.33\textwidth}
	\includegraphics[width=\linewidth, height=1.2\linewidth]{imgs/placeholder.jpeg}
\end{subfigure}
\begin{subfigure}[b]{.33\textwidth}
	\includegraphics[width=\linewidth, height=1.2\linewidth]{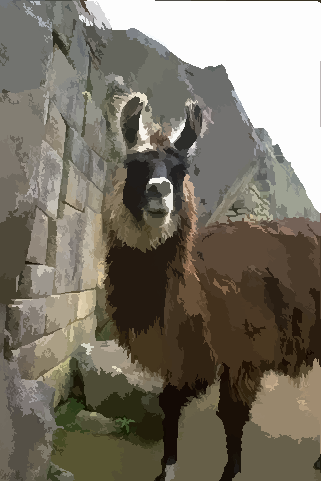}
	\caption{$K=1000$.}
\end{subfigure}
\begin{subfigure}[b]{.33\textwidth}
	\includegraphics[width=\linewidth, height=1.2\linewidth]{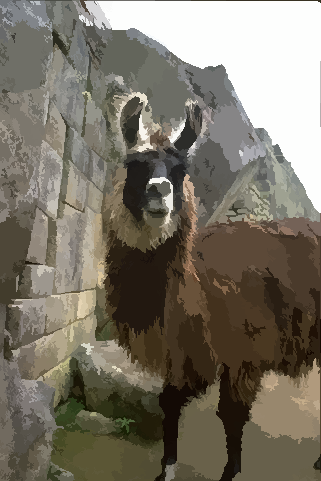}
	\caption{$K=1200$.}
\end{subfigure}
\caption{Segmentation results using our method on an image from the BSDS500 test set with varying number of superpixels. Each segmented image is represented with the average RGB pixel values of the corresponding superpixel.}
\label{fig:varying_k_1b}
\end{figure*}

In this section, we demonstrate the adaptiveness of our superpixels with the number of user-defined superpixel counts ranging from 200 to 1200.  
Figures~\ref{fig:varying_k_1a} and~\ref{fig:varying_k_1b} showcase the results in terms of superpixel boundaries and in terms of the average RGB pixel features per superpixel on an image from the BSDS500 test set. It can be observed easily from the view with superpixel boundaries (see Figure~\ref{fig:varying_k_1a}) that our technique gradually focuses on segmenting the texture-rich regions of the image as the user-specified number of superpixels increases. As a result, our superpixels are able to provide a very accurate and smooth representation of the original image, even with a relatively small number of superpixels (see Figure~\ref{fig:varying_k_1b}).

\begin{figure*}
	% line 1
	\begin{subfigure}[b]{.33\textwidth}
		\includegraphics[width=\linewidth, height=1.2\linewidth]{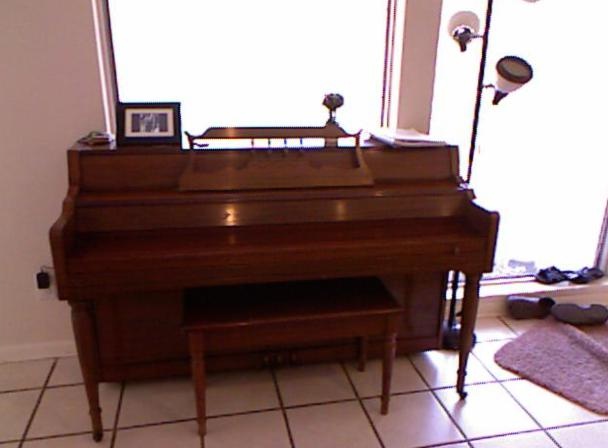}
		\caption{Original image.}
	\end{subfigure}
	\begin{subfigure}[b]{.33\textwidth}
		\includegraphics[width=\linewidth, height=1.2\linewidth]{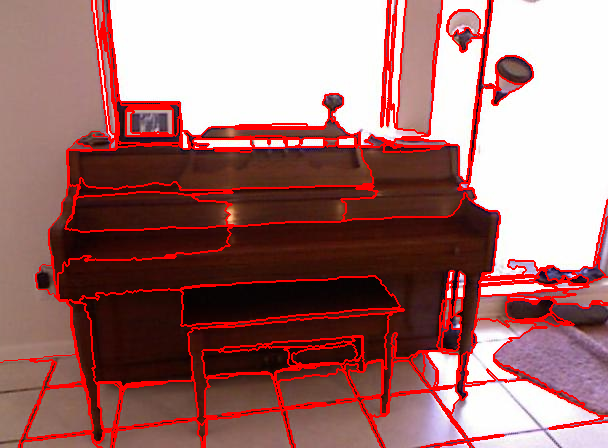}
		\caption{$K=200$.}
	\end{subfigure}
	\begin{subfigure}[b]{.33\textwidth}
		\includegraphics[width=\linewidth, height=1.2\linewidth]{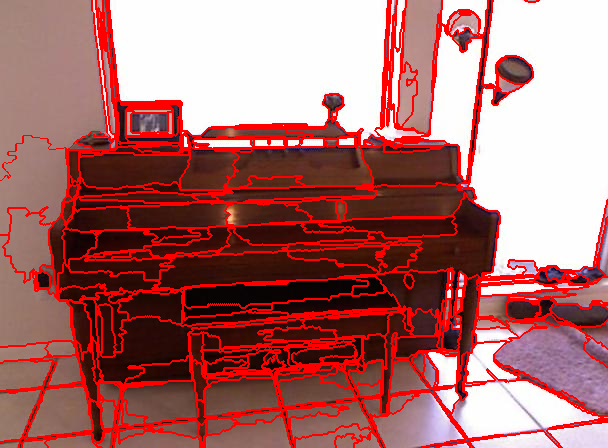}
		\caption{$K=400$.}
	\end{subfigure}
	% line 2
	\begin{subfigure}[b]{.33\textwidth}
		\includegraphics[width=\linewidth, height=1.2\linewidth]{imgs/placeholder.jpeg}
	\end{subfigure}
	\begin{subfigure}[b]{.33\textwidth}
		\includegraphics[width=\linewidth, height=1.2\linewidth]{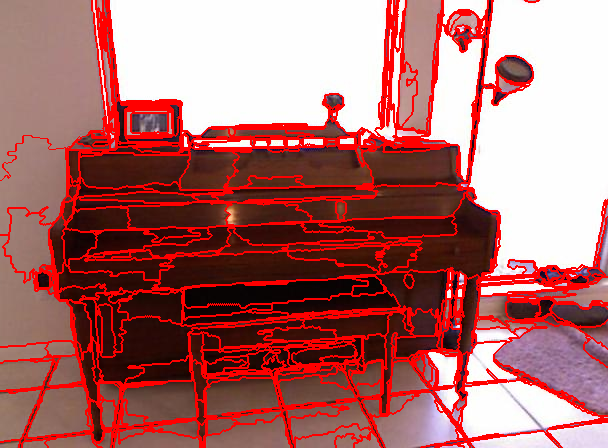}
		\caption{$K=600$.}
	\end{subfigure}
	\begin{subfigure}[b]{.33\textwidth}
		\includegraphics[width=\linewidth, height=1.2\linewidth]{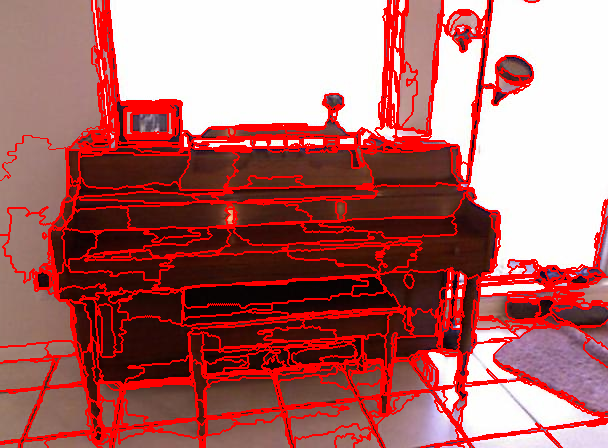}
		\caption{$K=800$.}
	\end{subfigure}
	% line 3
	\begin{subfigure}[b]{.33\textwidth}
		\includegraphics[width=\linewidth, height=1.2\linewidth]{imgs/placeholder.jpeg}
	\end{subfigure}
	\begin{subfigure}[b]{.33\textwidth}
		\includegraphics[width=\linewidth, height=1.2\linewidth]{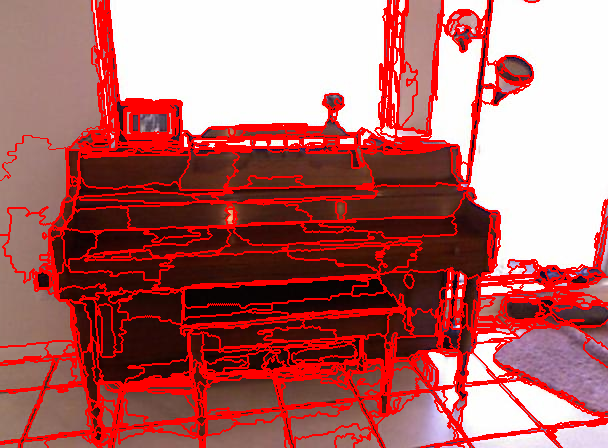}
		\caption{$K=1000$.}
	\end{subfigure}
	\begin{subfigure}[b]{.33\textwidth}
		\includegraphics[width=\linewidth, height=1.2\linewidth]{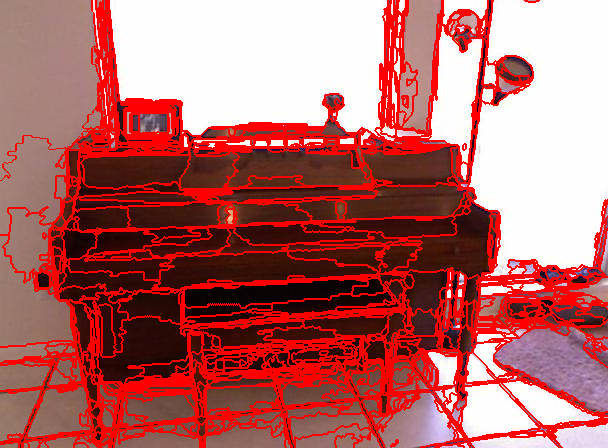}
		\caption{$K=1200$.}
	\end{subfigure}
	\caption{Segmentation results using our method on an image from the NYUv2 test set with varying number of superpixels.}
	\label{fig:varying_k_2a}
\end{figure*}

\begin{figure*}
	% line 1
	\begin{subfigure}[b]{.33\textwidth}
		\includegraphics[width=\linewidth, height=1.2\linewidth]{imgs/01410/01410.jpg}
		\caption{Original image.}
	\end{subfigure}
	\begin{subfigure}[b]{.33\textwidth}
		\includegraphics[width=\linewidth, height=1.2\linewidth]{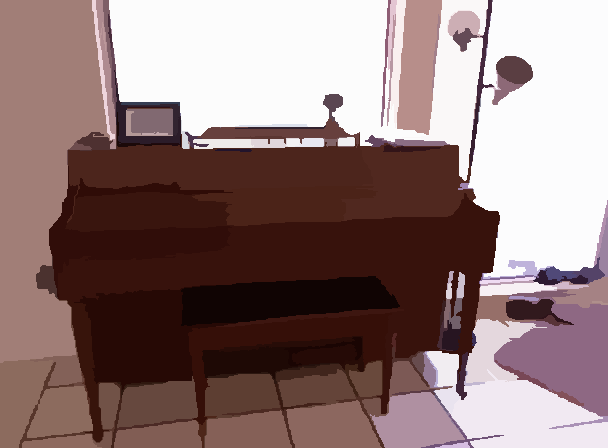}
		\caption{$K=200$.}
	\end{subfigure}
	\begin{subfigure}[b]{.33\textwidth}
		\includegraphics[width=\linewidth, height=1.2\linewidth]{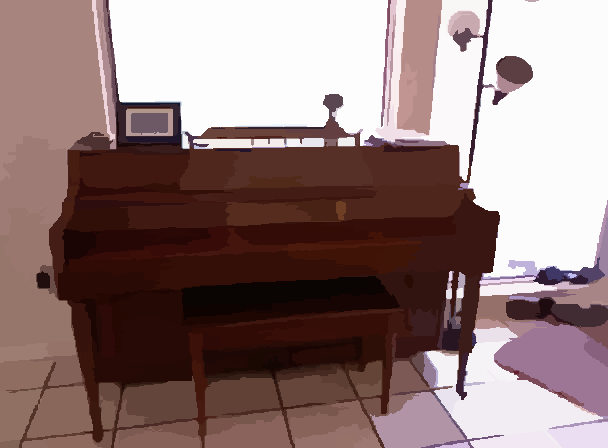}
		\caption{$K=400$.}
	\end{subfigure}
	% line 2
	\begin{subfigure}[b]{.33\textwidth}
		\includegraphics[width=\linewidth, height=1.2\linewidth]{imgs/placeholder.jpeg}
	\end{subfigure}
	\begin{subfigure}[b]{.33\textwidth}
		\includegraphics[width=\linewidth, height=1.2\linewidth]{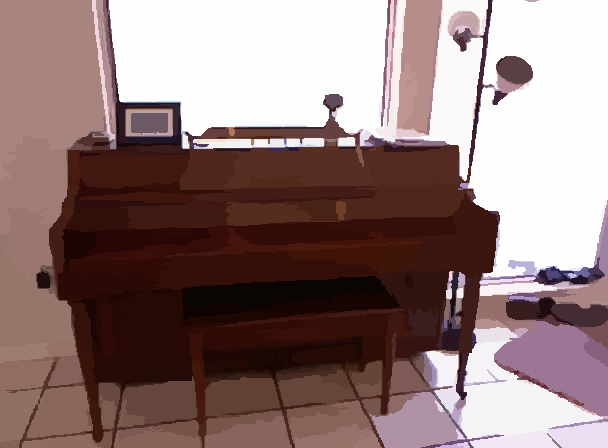}
		\caption{$K=600$.}
	\end{subfigure}
	\begin{subfigure}[b]{.33\textwidth}
		\includegraphics[width=\linewidth, height=1.2\linewidth]{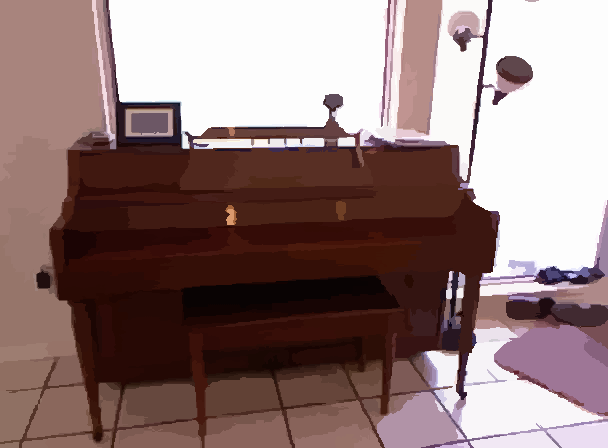}
		\caption{$K=800$.}
	\end{subfigure}
	% line 3
	\begin{subfigure}[b]{.33\textwidth}
		\includegraphics[width=\linewidth, height=1.2\linewidth]{imgs/placeholder.jpeg}
	\end{subfigure}
	\begin{subfigure}[b]{.33\textwidth}
		\includegraphics[width=\linewidth, height=1.2\linewidth]{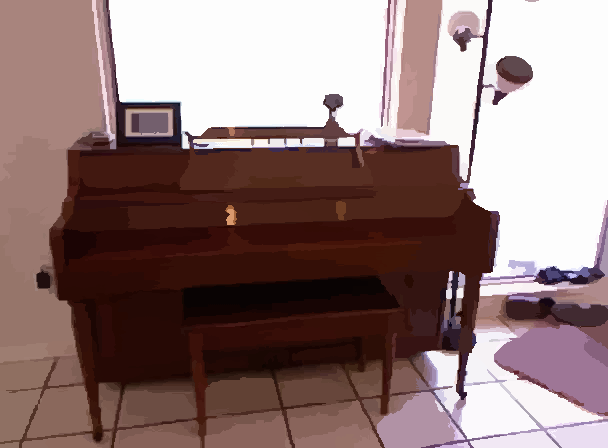}
		\caption{$K=1000$.}
	\end{subfigure}
	\begin{subfigure}[b]{.33\textwidth}
		\includegraphics[width=\linewidth, height=1.2\linewidth]{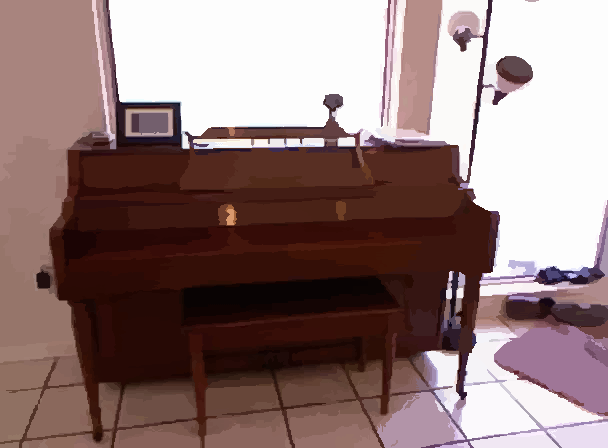}
		\caption{$K=1200$.}
	\end{subfigure}
	\caption{Segmentation results using our method on an image from the BSDS500 test set with varying number of superpixels. Each segmented image is represented with the average RGB pixel values of the corresponding superpixel.}
	\label{fig:varying_k_2b}
\end{figure*}

Similarly, the same benefits of our superpixels can be observed in indoor scenes in Figures~\ref{fig:varying_k_2a} and~\ref{fig:varying_k_2b}. We observe that our technique is able to delineate the main object boundaries in the image with 200 superpixels. With the increase of $K$, our technique further outlines the fine details within the identified objects. As a result, it is hard to even discern the difference visually between the original image (Figure~\ref{fig:varying_k_2b} (a)) and the superpixel representations of the image (see Figure~\ref{fig:varying_k_2b} (e) (f) (g)) at a first glance.

%% BSDS 1

\section{Superpixels for Saliency Detection}\label{sal_detect}
In this section, we present additional results on the application of our proposed DAL-HERS technique as a preprocessing tool for the downstream task of saliency detection~\cite{hou2007saliency, goferman2011context, zhu2014saliency}. 
%% Explain what saliency detection is 
The main purpose of this task is to extract the most salient object from its background in an image.
%, which can then be used for other computer vision applications, such as video object tracking. 
%% Explain the saliency detection method that we use  
One of the most classical techniques in saliency detection is Saliency Optimisation (SO)~\cite{zhu2014saliency}. SO first segments an image into a number of superpixels, and then constructs an undirected weighted graph using the superpixels as primitives for further detecting the salient regions among them. Here, the superpixels are produced using SLIC due to its simplicity and efficiency.

\begin{figure*}[ht!]
	\centering
	%%%%%%%%%%%%%%%%%%%%%%%%%%%%%%%%%%%%%%%%
	\begin{subfigure}[b]{.24\textwidth}
		%% original 
		\includegraphics[width=\linewidth, height=.7\linewidth]{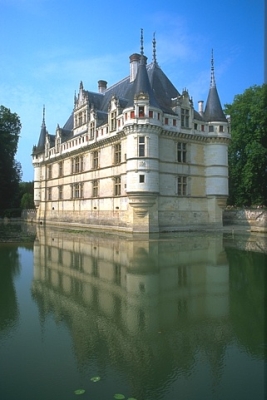}
		%\caption{Original.}
	\end{subfigure}
	\begin{subfigure}[b]{.24\textwidth}
		\includegraphics[width=\linewidth, height=.7\linewidth]{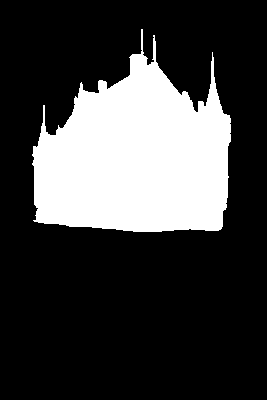}
		%\caption{Ground truth.}
	\end{subfigure}
	\begin{subfigure}[b]{.24\textwidth}
		%% SLIC
		\includegraphics[width=\linewidth, height=.7\linewidth]{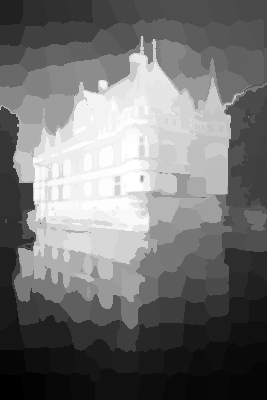}
	%	\caption{SLIC.}
	\end{subfigure}
	\begin{subfigure}[b]{.24\textwidth}
		%% ours 
		\includegraphics[width=\linewidth, height=.7\linewidth]{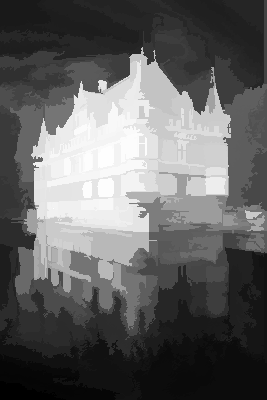}
	%	\caption{Ours.}
	\end{subfigure}
	\vfill 
	%%%%%%%%%%%%%%%%%%%%%%%%%%%%%%%%%%%%%%%%
	\begin{subfigure}[b]{.24\textwidth}
		%% original 
		\includegraphics[width=\linewidth, height=.7\linewidth]{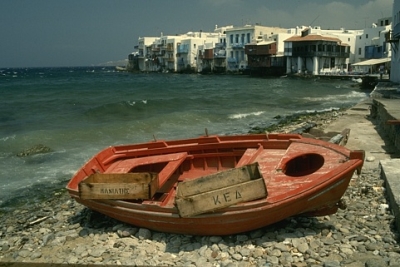}
	%	\caption{Original.}
	\end{subfigure}
	\begin{subfigure}[b]{.24\textwidth}
		\includegraphics[width=\linewidth, height=.7\linewidth]{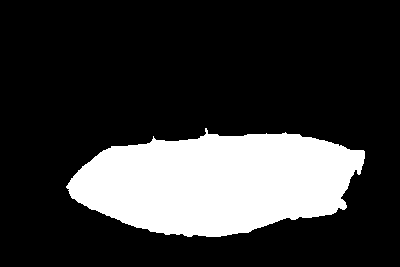}
	%	\caption{Ground truth.}
	\end{subfigure}
	\begin{subfigure}[b]{.24\textwidth}
		%% SLIC
		\includegraphics[width=\linewidth, height=.7\linewidth]{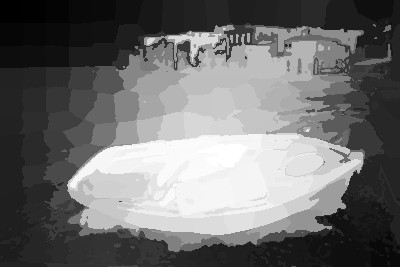}
	%	\caption{SLIC.}
	\end{subfigure}
	\begin{subfigure}[b]{.24\textwidth}
		%% ours 
		\includegraphics[width=\linewidth, height=.7\linewidth]{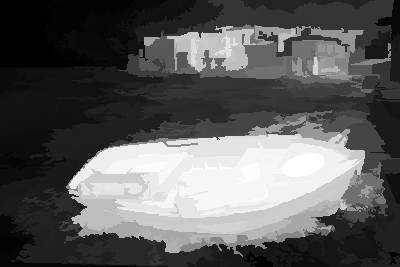}
	%	\caption{Ours.}
	\end{subfigure}
	\vfill
	%%%%%%%%%%%%%%%%%%%%%%%%%%%%%%%%%%%%%%%%
	\begin{subfigure}[b]{.24\textwidth}
		%% original 
		\includegraphics[width=\linewidth, height=.7\linewidth]{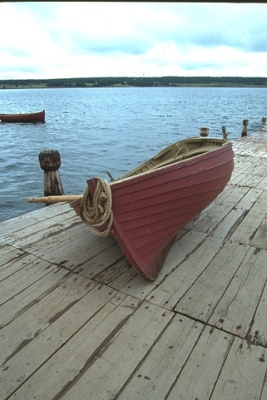}
		%	\caption{Original.}
	\end{subfigure}
	\begin{subfigure}[b]{.24\textwidth}
		\includegraphics[width=\linewidth, height=.7\linewidth]{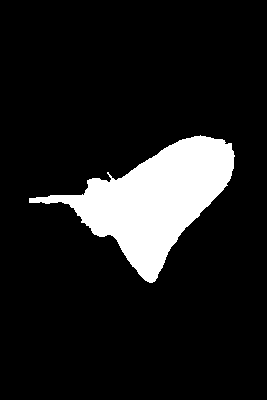}
		%	\caption{Ground truth.}
	\end{subfigure}
	\begin{subfigure}[b]{.24\textwidth}
		%% SLIC
		\includegraphics[width=\linewidth, height=.7\linewidth]{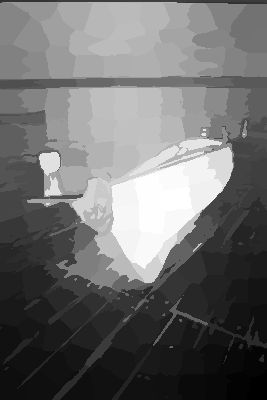}
		%	\caption{SLIC.}
	\end{subfigure}
	\begin{subfigure}[b]{.24\textwidth}
		%% ours 
		\includegraphics[width=\linewidth, height=.7\linewidth]{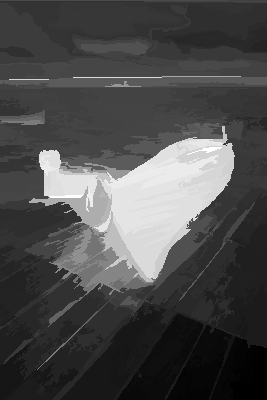}
		%	\caption{Ours.}
	\end{subfigure}
	\vfill
	%%%%%%%%%%%%%%%%%%%%%%%%%%%%%%%%%%%%%%%%
	\begin{subfigure}[b]{.24\textwidth}
		%% original 
		\includegraphics[width=\linewidth, height=.7\linewidth]{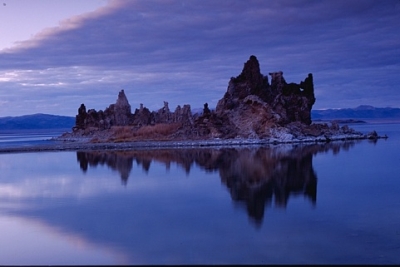}
		%	\caption{Original.}
	\end{subfigure}
	\begin{subfigure}[b]{.24\textwidth}
		\includegraphics[width=\linewidth, height=.7\linewidth]{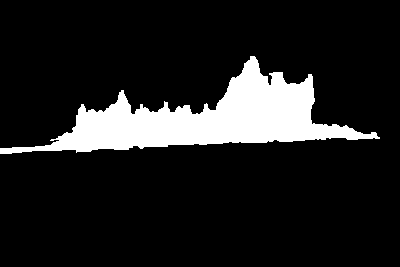}
		%	\caption{Ground truth.}
	\end{subfigure}
	\begin{subfigure}[b]{.24\textwidth}
		%% SLIC
		\includegraphics[width=\linewidth, height=.7\linewidth]{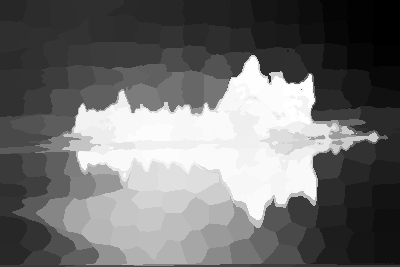}
		%	\caption{SLIC.}
	\end{subfigure}
	\begin{subfigure}[b]{.24\textwidth}
		%% ours 
		\includegraphics[width=\linewidth, height=.7\linewidth]{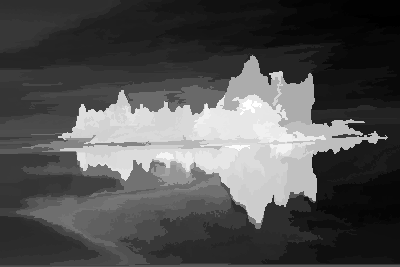}
		%	\caption{Ours.}
	\end{subfigure}
	\vfill
	%%%%%%%%%%%%%%%%%%%%%%%%%%%%%%%%%%%%%%%%
	\begin{subfigure}[b]{.24\textwidth}
		%% original 
		\includegraphics[width=\linewidth, height=.7\linewidth]{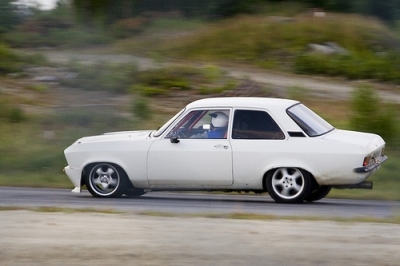}
		%	\caption{Original.}
	\end{subfigure}
	\begin{subfigure}[b]{.24\textwidth}
		\includegraphics[width=\linewidth, height=.7\linewidth]{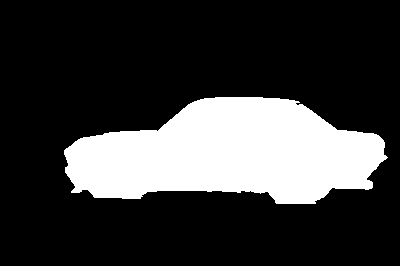}
		%	\caption{Ground truth.}
	\end{subfigure}
	\begin{subfigure}[b]{.24\textwidth}
		%% SLIC
		\includegraphics[width=\linewidth, height=.7\linewidth]{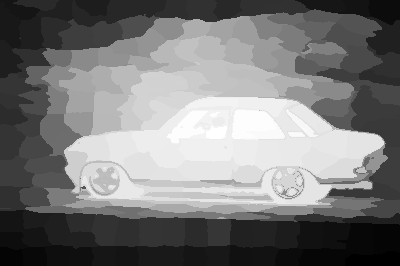}
		%	\caption{SLIC.}
	\end{subfigure}
	\begin{subfigure}[b]{.24\textwidth}
		%% ours 
		\includegraphics[width=\linewidth, height=.7\linewidth]{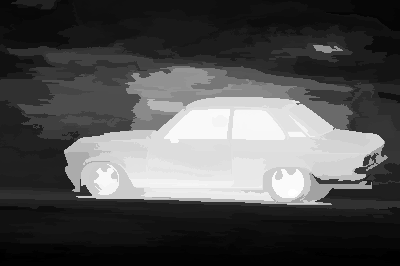}
		%	\caption{Ours.}
	\end{subfigure}
	\vfill
	%%%%%%%%%%%%%%%%%%%%%%%%%%%%%%%%%%%%%%%%
	\begin{subfigure}[b]{.24\textwidth}
		%% original 
		\includegraphics[width=\linewidth, height=.7\linewidth]{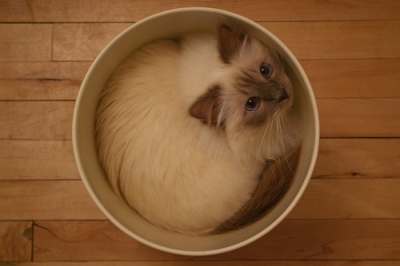}
	%	\caption{Original.}
	\end{subfigure}
	\begin{subfigure}[b]{.24\textwidth}
		\includegraphics[width=\linewidth, height=.7\linewidth]{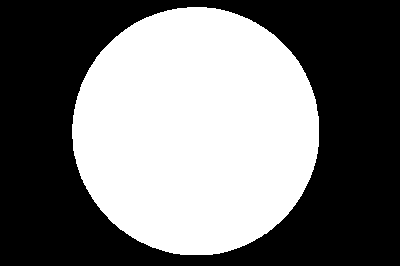}
	%	\caption{Ground truth.}
	\end{subfigure}
	\begin{subfigure}[b]{.24\textwidth}
		%% SLIC
		\includegraphics[width=\linewidth, height=.7\linewidth]{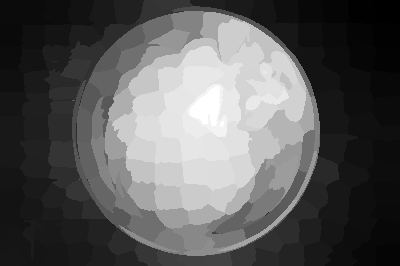}
	%	\caption{SLIC.}
	\end{subfigure}
	\begin{subfigure}[b]{.24\textwidth}
		%% ours 
		\includegraphics[width=\linewidth, height=.7\linewidth]{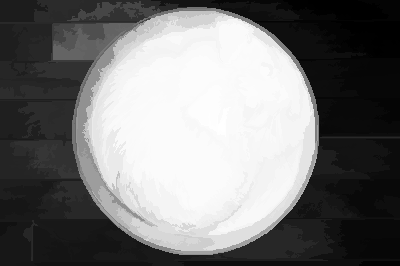}
	%	\caption{Ours.}
	\end{subfigure}
	\vfill 
	%%%%%%%%%%%%%%%%%%%%%%%%%%%%%%%%%%%%%%%%
	\begin{subfigure}[b]{.24\textwidth}
		%% original 
		\includegraphics[width=\linewidth, height=.7\linewidth]{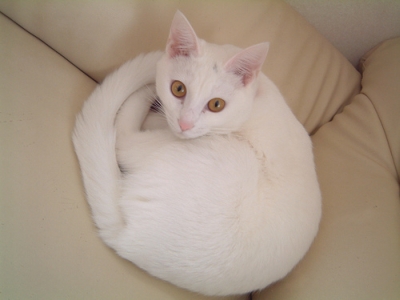}
		%	\caption{Original.}
	\end{subfigure}
	\begin{subfigure}[b]{.24\textwidth}
		\includegraphics[width=\linewidth, height=.7\linewidth]{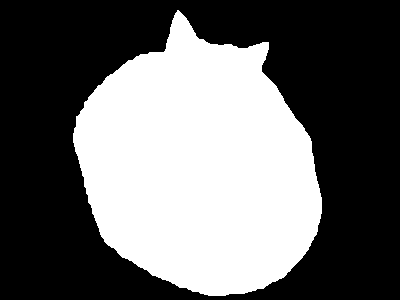}
		%	\caption{Ground truth.}
	\end{subfigure}
	\begin{subfigure}[b]{.24\textwidth}
		%% SLIC
		\includegraphics[width=\linewidth, height=.7\linewidth]{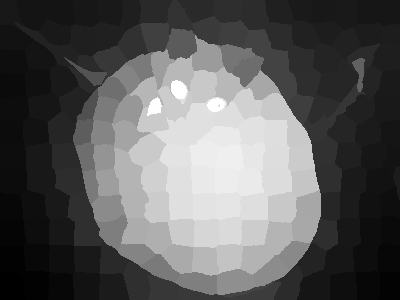}
		%	\caption{SLIC.}
	\end{subfigure}
	\begin{subfigure}[b]{.24\textwidth}
		%% ours 
		\includegraphics[width=\linewidth, height=.7\linewidth]{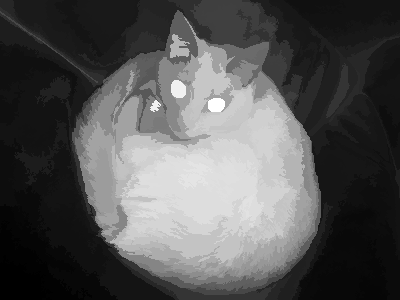}
		%	\caption{Ours.}
	\end{subfigure}
	\vfill 
	%%%%%%%%%%%%%%%%%%%%%%%%%%%%%%%%%%%%%%%%
	\begin{subfigure}[b]{.24\textwidth}
		%% original 
		\includegraphics[width=\linewidth, height=.7\linewidth]{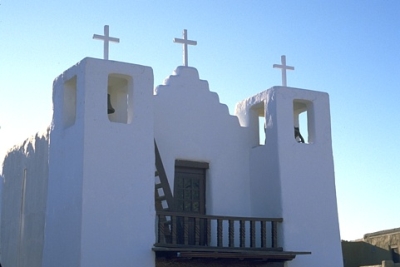}
		%	\caption{Original.}
	\end{subfigure}
	\begin{subfigure}[b]{.24\textwidth}
		\includegraphics[width=\linewidth, height=.7\linewidth]{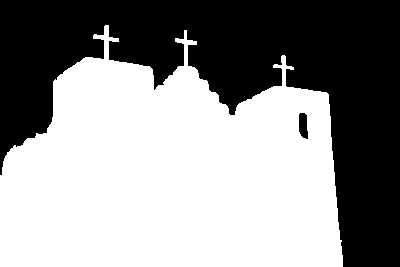}
		%	\caption{Ground truth.}
	\end{subfigure}
	\begin{subfigure}[b]{.24\textwidth}
		%% SLIC
		\includegraphics[width=\linewidth, height=.7\linewidth]{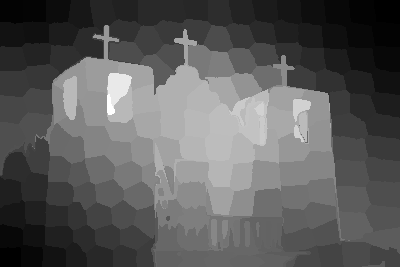}
		%	\caption{SLIC.}
	\end{subfigure}
	\begin{subfigure}[b]{.24\textwidth}
		%% ours 
		\includegraphics[width=\linewidth, height=.7\linewidth]{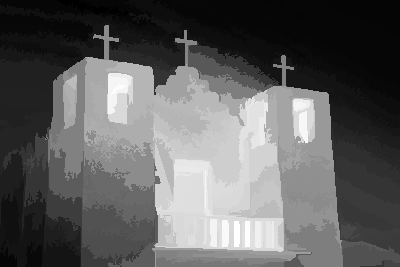}
		%	\caption{Ours.}
	\end{subfigure}
	\vfill 
	\caption{Saliency detection results on sample images from the ECSSD dataset~\cite{shi2015hierarchical} with 200 superpixels. From left to right: original image, ground truth segmentation mask, SLIC, DAL-HERS.}
	\label{fig:sd}
\end{figure*}

%% Explain the experiments we conducted (the SD method, the performance measures)
To demonstrate the advantages of our proposed DAL-HERS technique, we replace SLIC with DAL-HERS in the saliency detection process. We report both the quantitative comparison in Table~\ref{tab_sd} and the visual comparisons in Figure~\ref{fig:sd} in terms of the standard performance metrics on the ECSSD dataset~\cite{shi2015hierarchical}. 
%% Comment on the qualitative results 
It can be seen from Table~\ref{tab_sd} that our DAL-HERS technique enjoys an obvious advantage over SLIC across all three metrics. This advantage is further supported by the visual comparisons in Figure~\ref{fig:sd}. It is clear that the results produced using SLIC are very non-smooth and segmented, which is due to the non-adaptive nature of SLIC superpixels. Whereas the results produced with DAL-HERS are smooth whilst highlighting the contours of the most salient object in an image.
\begin{table}[htbp!]
	\centering
	\begin{tabular}{|c|c|c|c|}
		\hline
		Method & $F$-measure & weighted $F_{\beta}$ & MAE \\ 
		\hline 
		SLIC &0.885 &0.432&0.286\\
		\hline
		DAL-HERS &\textbf{0.906}&\textbf{0.520}&\textbf{0.200}\\
		\hline 
	\end{tabular}
	\caption{Superpixels for saliency object detection.}
	\label{tab_sd}
\end{table}

%{\small
%\bibliographystyle{ieee_fullname}
%\bibliography{supp}
%}